\documentclass[10pt,twocolumn,letterpaper]{article}

\usepackage[pagenumbers]{iccv} %

\usepackage{colortbl}
\definecolor{iccvblue}{rgb}{0.21,0.49,0.74}
\renewcommand{\thefootnote}{\fnsymbol{footnote}}
\newcommand*\samethanks[1][\value{footnote}]{\footnotemark[#1]}
\usepackage[pagebackref,breaklinks,colorlinks,allcolors=iccvblue]{hyperref}

\def\paperID{108} %
\def\confName{ICCV}
\def\confYear{2025}

\title{Emergent Active Perception and Dexterity of Simulated Humanoids \\ from Visual Reinforcement Learning }

\author{
Zhengyi Luo$^{1,2}$ \thanks{\tiny{Equal Contribution}}
\quad
Chen Tessler$^{1,2}$ \samethanks
\quad
Toru Lin$^{1,3}$
\quad
Ye Yuan$^{1}$
\quad
Tairan He$^{1,2}$
\quad
Wenli Xiao$^{1,2}$ \\
\quad
Yunrong Guo$^{1}$ 
\quad
Gal Chechik$^{1}$
\quad
Kris Kitani$^{2}$
\quad
Linxi  Fan$^{1}$ \thanks{\tiny{Equal Advising}}
\quad
Yuke Zhu$^{1}$ \samethanks
\\
$^{1}$Nvidia; $^{2}$Carnegie Mellon University; $^{3}$University of California, Berkeley\\
{\tt\small \href{\link}{\link}  }}

\usepackage{amsfonts}
\usepackage{bm}
\usepackage{amsmath}
\usepackage{mathtools}
\usepackage{multirow}
\usepackage{booktabs}
\usepackage{caption}
\usepackage{enumitem}
\usepackage{wrapfig,lipsum,booktabs}
\usepackage{caption}
\usepackage{bbm}
\usepackage{multirow}
\usepackage{pifont}
\usepackage{subcaption}
\usepackage{xcolor}
\usepackage[linesnumbered,ruled,vlined]{algorithm2e}

\SetCommentSty{mycommfont}
\SetAlCapFnt{\small} 
\usepackage{etoc}
\SetAlgorithmName{Algo}{Algo}{}

\definecolor{verylightgray}{RGB}{240, 240, 240}
\definecolor{llnvgreen}{RGB}{220, 237, 191}

\definecolor{rqred}{RGB}{247, 202, 201}
\definecolor{rqblue}{RGB}{230,230,250}

\definecolor{gsred}{RGB}{201, 201, 247}

\definecolor{mydarkblue}{rgb}{0,0.08,0.45}
\definecolor{mydarkgreen}{RGB}{0, 139, 69}
\definecolor{mygreen2}{RGB}{0 205 0}
\definecolor{mybrown}{RGB}{139 69 19}
\definecolor{Methodred}{RGB}{191, 3, 3}

\renewcommand{\paragraph}[1]{{\vspace{1mm}\noindent \bf #1}.}

\newcommand{\reals}{\mathbb{R}}

\newcommand{\encoder}{\bs{\mathcal{E}}_{\text{PULSE-X}}}
\newcommand{\decoder}{\bs{\mathcal{D}}_{\text{PULSE-X}}}
\newcommand{\prior}{\bs{\mathcal{P}}_{\text{PULSE-X}}}

\newcommand{\latentt}{\bs{z}_t}

\newcommand{\imaget}{\bs{I}_{t}}

\newcommand{\priormu}{\bs{\mu}^{p}_t}
\newcommand{\priorsigma}{\bs{\sigma}^{p}_t}
\newcommand{\encodermu}{\bs{\mu}^{e}_t}

\newcommand{\encodersigma}{\bs{\sigma}^{e}_t}

\newcommand{\link}{https://zhengyiluo.github.io/PDC}

\newcommand{\name}{\text{PDC}\xspace}
\newcommand{\namepulsex}{\text{PULSE-X}\xspace}
\newcommand{\namephcx}{\text{PHC-X}\xspace}

\newcommand{\policyours}{{\pi_{\text{\name}}}}

\newcommand{\policy}{{\pi_{\text{PDC}}}}
\newcommand{\rewardfunc}{\bs{\mathcal{R}}}

\newcommand{\rewardapproach}{r^{\text{approach}}_t}
\newcommand{\rewardpregrasp}{r^{\text{pre-grasp}}_t}
\newcommand{\rewardobj}{r^{\text{obj}}_t}
\newcommand{\rewarddrawer}{r^{\text{drawer}}_t}
\newcommand{\rewardlookat}{r^{\text{lookat}}_t}
\newcommand{\rewardlookatobject}{r^{\text{lookat-object}}_t}
\newcommand{\rewardlookatmarker}{r^{\text{lookat-marker}}_t}
\newcommand{\gazevector}{\mathbf{v}_{\text{gaze}}}
\newcommand{\targetvector}{\mathbf{v}_{\text{eye-target}}}

\newcommand{\refqpregrasp}{{\bs{\hat{q}}^{\text{pre-grasp}}}}
\newcommand{\reftpregrasp}{{\bs{\hat{p}}^{\text{pre-grasp}}}}
\newcommand{\refrpregrasp}{{\bs{\hat{\theta}}^{\text{pre-grasp}}}}

\newcommand{\simr}{{\bs{{\theta}}_{t}}}
\newcommand{\simt}{{\bs{{p}}_{t}}}
\newcommand{\simthand}{{\bs{{p}}_{t}^{\text{H}}}}
\newcommand{\contacthand}{{\bs{{c}}_{t}}}
\newcommand{\simrhand}{{\bs{{\theta}}_{t}^{\text{hand}}}}
\newcommand{\simtphand}{{\bs{{p}}_{t-1}^{\text{hand}}}}
\newcommand{\goalstate}{{\bs{o}^{\text{g}}_t}}

\newcommand{\goalstateimitate}{{\bs{o}^{\text{g-mimic}}_t}}

\newcommand{\latentttask}{\bs{z}^{\text{\name}}_t}

\newcommand{\selfstate}{{\bs{o}^{\text{p}}_t}}

\newcommand{\state}{{\bs{s}_t}}
\newcommand{\staten}{{\bs{s}_{t+1}}}
\newcommand{\obs}{{\bs{o}_t}}
\newcommand{\obsomnigrasp}{{\bs{o}^{\text{Omnigrasp}}_t}}
\newcommand{\obsim}{{\bs{o}^{\text{imitation}}_t}}
\newcommand{\action}{{\bs{a}_t}}

\newcommand{\mpjpe}{E_\text{mpjpe}}

\newcommand{\gmpjpe}{E_\text{g-mpjpe}}

\newcommand{\acc}{E_\text{acc}}
\newcommand{\vel}{E_\text{vel}}
\newcommand{\success}{\text{Succ}}
\newcommand{\poserror}{E_\text{pos}}

\newcommand{\successsearch}{\text{Succ}_{\text{search}}}
\newcommand{\successgrasp}{\text{Succ}_{\text{grasp}}}
\newcommand{\successdrawer}{\text{Succ}_{\text{drawer}}}
\newcommand{\successtraj}{\text{Succ}_{\text{traj}}}
\newcommand{\successleft}{\text{Succ}_{\text{left}}}
\newcommand{\successright}{\text{Succ}_{\text{right}}}
\newcommand{\successbi}{\text{Succ}_{\text{bi}}}

\newcommand{\objr}{\bs{\theta}^{\text{obj}}_{t}}
\newcommand{\objt}{\bs{p}^{\text{obj}}_{t}}

\newcommand{\targetobj}{\bs{p}^{\text{target}}_{t}}

\newcommand{\refobjt}{\bs{\hat{ p}}^{\text{obj}}_{t}}

\newcommand{\refobjtn}{\bs{\hat{ p}}^{\text{obj}}_{t+1}}

\newcommand{\handinfo}{\bs{h}_{t}}
\newcommand{\handinfoleft}{\bs{h}^{\text{left}}_{t}}
\newcommand{\handinforight}{\bs{h}^{\text{right}}_{t}}
\newcommand{\handinfotime}{\bs{h}^{\text{time}}_{t}}

\newcommand{\objlt}{\bs{{ \sigma}}^{\text{obj}}}

\newcommand{\simav}{{\bs{{\omega}}_{t}}}
\newcommand{\simlv}{{\bs{v}_{t}}}

\newcommand{\contactstart}{\lambda_{\text{start}}}
\newcommand{\contactend}{\lambda_{\text{end}}}

\newcommand{\bs}[1]{\boldsymbol{#1}}
\newcommand{\cmark}{\ding{51}}%
\newcommand{\xmark}{\ding{55}}%

\newcommand\blfootnote[1]{%
  \begingroup
  \renewcommand\thefootnote{}\footnote{#1}%
  \addtocounter{footnote}{-1}%
  \endgroup
}

\begin{document}

\Crefname{equation}{Eq.}{Eqs.}
\Crefname{figure}{Fig.}{Figs.}
\Crefname{tabular}{Tab.}{Tabs.}
\Crefname{appendix}{Appendix.}{Appendix.}

\etocdepthtag.toc{mtchapter}
\etocsettagdepth{mtchapter}{subsection}
\etocsettagdepth{mtappendix}{none}

\twocolumn[{
\renewcommand\twocolumn[1][]{#1}%
\maketitle
\vspace{-0.4in}
\begin{center}
    \centering
    \includegraphics[width=1\textwidth]{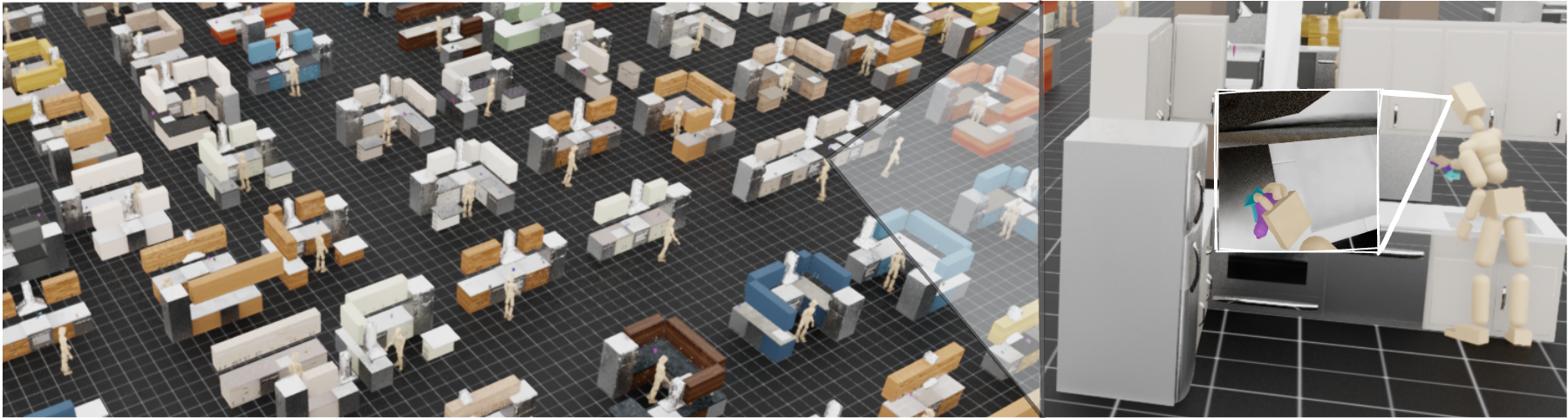}
    \vspace{-7mm}
    
    \captionof{figure}{\small{Perceptive Dexterous Control (\name) enables a humanoid equipped with egocentric vision to search for, reach, grasp, and manipulate objects in cluttered kitchen scenes. We use visual perception as the sole interface for indicating which hand to use, which object to grasp, where to move, and which drawer to open.}}
    \label{fig:teaser}
\end{center}%
}]
\maketitle

\begin{abstract}

Human behavior is fundamentally shaped by visual perception --- our ability to interact with the world depends on actively gathering relevant information and adapting our movements accordingly. Behaviors like searching for objects, reaching, and hand-eye coordination naturally emerge from the structure of our sensory system. Inspired by these principles, we introduce Perceptive Dexterous Control (\name), a framework for vision-driven dexterous whole-body control with simulated humanoids. \name operates solely on egocentric vision for task specification, enabling object search, target placement, and skill selection through visual cues, without relying on privileged state information (\eg, 3D object positions and geometries). This perception-as-interface paradigm enables learning a single policy to perform multiple household tasks, including reaching, grasping, placing, and articulated object manipulation. We also show that training from scratch with reinforcement learning can produce emergent behaviors such as active search. These results demonstrate how vision-driven control and complex tasks induce human-like behaviors and can serve as the key ingredients in closing the perception-action loop for animation, robotics, and embodied AI.

\end{abstract}
    
\blfootnote{$^*$Equal contribution, $\dag$ GEAR Team Leads}
\section{Introduction}
\label{sec:intro}

Human interaction with the world is fundamentally governed by active perception --- a continuous process where vision, touch, and proprioception dynamically guide our movements and interactions~\cite{o2001sensorimotor, noe2004action}.  In this work, we study how a simulated humanoid, equipped only with egocentric perception, can perform tasks such as object transportation and articulated object manipulation in diverse household scenes. The combination of high-dimensional visual input and high-degree-of-freedom humanoid control makes this problem computationally demanding. As a result, humanoid controllers in animation and robotics often rely on privileged state information (\eg, precise 3D object pose and shape), sidestepping the challenges of processing noisy visual input. This raises a key question: How can we effectively close the egocentric perception-action loop to enable humanoids to interact with their environment using only their egocentric sensory capabilities?

Perception-driven humanoid control presents unique challenges because it involves two computationally intensive problems: dexterous whole-body control and visual perception in natural environments. While loco-manipulation demands precise coordination of balance, limb movements, and dexterity, incorporating vision introduces additional complexities of partial observability and the need for active information gathering. Due to these challenges, prior work \cite{luoomnigrasp, braun2023physically, wang2023physhoi} has relied mainly on privileged object-state information to achieve humanoid-object interaction, bypassing the difficulties of noisy sensory input and gaze control.
While using privileged information provides a compact, pre-processed representation of the environment that includes objects outside the field of view, it fundamentally limits the emergence of human-like behaviors such as visual search since the humanoid has complete knowledge of object locations at all times. 
In contrast, learning from visual observations eliminates this limitation but introduces several key challenges: the policy must learn to handle partial observability and tackle the increased input dimensionality (such as 128$\times$128$\times$3 RGB images) --- which is orders of magnitude larger than state representations. These challenges exacerbate the significant sample efficiency issues in reinforcement learning (RL) for whole-body humanoid control.

Another challenge in developing vision-driven humanoids is designing an input representation that facilitates learning multiple tasks simultaneously. Consider a kitchen cleaning scenario: the humanoid must identify specific objects in cluttered environments, understand task commands, and precisely place items in designated locations. Each of these steps requires clear task instructions. While natural language offers flexibility as an interface, it often lacks the precision required for exact object selection and placement in physical space. Previous approaches, such as Catch \& Carry~\cite{Merel2020-qm}, have combined vision-based policies with phase variables, but this design limits scalability since each new task requires retraining the policy to handle new task variables. The ideal input representation should be task-agnostic, allowing the policy to learn in diverse manipulation scenarios without task-specific modifications.

To address these challenges, we introduce Perceptive Dexterous Control (\name), a framework for learning human-like whole-body behaviors of simulated humanoids through perception-driven control. \name controls humanoids to act solely from visual input (RGB, RGB-D, or Stereo) and proprioception. Rather than using state variables (\eg, task phases, object information) to encode tasks, \name specifies tasks directly through perception. Similarly to how augmented reality enhances human vision via informative markers and visual cues, we enhance the robot's vision using object masks and 3D markers to select objects and specify target locations. This vision-centric approach supports continuously adapting the agent's skills to new tasks and complex scenes. To tackle the challenging whole-body dexterous control task, we leverage control priors learned from large-scale motion capture data. Combined with household settings and dexterous hand tasks, these priors enable RL policies to develop emergent human-like behaviors --- including active search and whole-body coordination.

To summarize, our contributions are as follows: 
(1) We demonstrate the feasibility of vision-driven, whole-body dexterous humanoid control in naturalistic household environments, a novel and challenging task that involves both perception and complex motor control;
(2) We introduce a vision-driven task specification paradigm that uses visual cues for object selection, target placement, and skill specification, eliminating the need for predefined state variables;
(3) We show that human-like behaviors, including active search and whole-body coordination, emerge from our vision-driven approach. Extensive evaluation shows how different visual modalities (RGB, RGBD, Stereo) and reward designs impact behaviors and task performance (\eg, stereo outperforms RGB by 9\% in success rate).

\begin{figure*}[t!]
    \centering
    \includegraphics[width=0.95\textwidth]{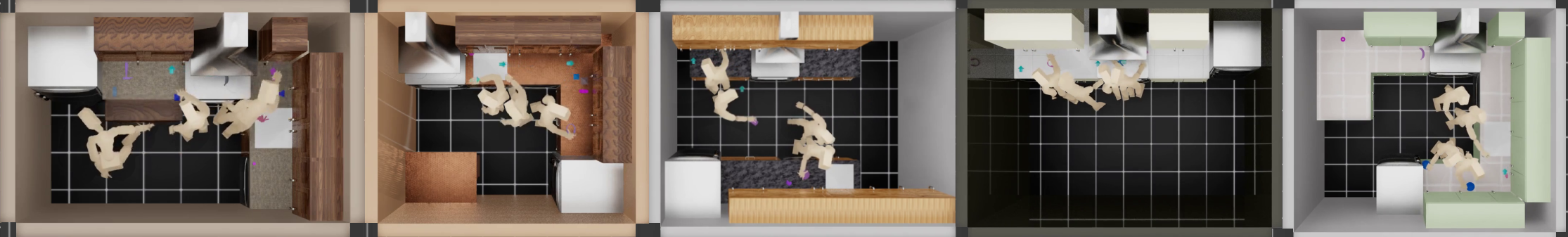}
    \caption{\small  \textbf{Kitchens:} Our agent is trained in parallel on a large set of (randomly) procedurally generated kitchens. Each generated kitchen is structurally and visually different. Objects are spawned in random locations on the counter, and the agent starts in a random position and orientation within the scene. This diversity encourages the agent to learn general behaviors, such as search, and robust interaction.}
    \label{fig:kitchens}
    \vspace{-0.4cm}
\end{figure*}

\section{Related Work}
\label{sec:related_works}

\paragraph{Visual Dexterous Manipulation} Learning visual dexterous policies for simulated \cite{xu2023unidexgrasp,wan2023unidexgrasp++} and real-world \cite{lin2024learning, lin2024twisting, singh2024hand, qi2023general, chen2023visual, lin2025sim, xu2023dexterous, singh2024dextrah, lum2024dextrah, chen2024vividex, qin2022dexmv, cheng2024open, qin2023anyteleop} dexterous hands has gained increasing attention due to its relevance to real-world tasks such as object picking and manipulation. These approaches can be roughly divided into RL \cite{lin2025sim, lum2024dextrah, singh2024dextrah, xu2023unidexgrasp} and behavior cloning \cite{cheng2024open, qin2023anyteleop, lin2024learning} methods. \name falls under RL, where the dexterous hands' behavior is learned through trial and error. Among RL methods, a popular paradigm is to first learn a state-space policy using privileged object information from simulation and then distill it into vision-based policies \cite{xu2023unidexgrasp,wan2023unidexgrasp++, lum2024dextrah, singh2024dextrah}. However, to successfully distill, these methods assume a fixed camera pointing at a stationary table and do not involve active perception. For mobile and dexterous humanoids that loco-manipulate, where to look and how to pick up objects remain open questions.

\paragraph{Whole Body Loco-Manipulation} Controlling simulated \cite{luoomnigrasp, bae2023pmp, braun2023physically, xu2025intermimicuniversalwholebodycontrol, Merel2020-qm, xie2023hierarchical, xu2025humanvla, tirumala2024learning, Haarnoja2023-qz, wang2023physhoi} and real humanoids \cite{lu2024mobile, he2024omnih2o} to loco-manipulate requires precise coordination of the lower and upper body. Some prior work adopts a reference motion imitation approach \cite{xu2025intermimicuniversalwholebodycontrol, wang2023physhoi, he2024omnih2o, lu2024mobile}, where the humanoid learns from reference human-object interactions. Others follow a generative approach \cite{luoomnigrasp, bae2023pmp, xie2023hierarchical, xu2025humanvla, tirumala2024learning, Haarnoja2023-qz}, where the humanoid is provided high-level objectives (such as moving boxes \cite{tirumala2024learning}, grasping objects \cite{luoomnigrasp}, \etc) to complete but not a reference motion to follow.  Closely related to ours, Omnigrasp \cite{luoomnigrasp} studies a state-space policy to grasp diverse objects using either hand (predominantly the right hand). HumanVLA \cite{xu2025humanvla} learns a language-conditioned visual policy for room arrangement via distillation. Compared to \name, HumanVLA does not use dexterous hands and always assumes the object of interest and target location are within view of the humanoid during initialization. Similarly to our setting of egocentric vision, Catch \& Carry \cite{Merel2020-qm} learns a vision-conditioned policy for carrying boxes and catching balls and uses phase variables and privileged information (object position for carrying). Compared to Catch \& Carry, we involve dexterous hands, learn multiple tasks, and do not use privileged information.

\paragraph{Perception-as-Interface} The interface theorem of perception \cite{hoffman2015interface} argues that perception functions as an interface for guiding useful actions instead of reconstructing the 3D world. Recently, many \cite{huang2024rekep, xu2024flow, nasiriany2024pivot} have explored using visual signals as indicators for policies to imbue \cite{nasiriany2024pivot} common sense or guide actions \cite{xu2024flow}. In games and augmented reality (AR), 3D markers and visual signals are also constantly used to indicate actions or give instructions. Building upon this idea, we leverage the redundant nature of images and use visual cues to replace state input for our RL policy.

\paragraph{Hierarchical Reinforcement Learning} %
The options framework, also known as skills \cite{sutton1999between}, provides an abstraction layer for RL agents, enabling the separation between high-level planning and low-level execution \cite{Peng2017-il, Merel2018-ah, haarnoja2018latent, tessler2017deep, lum2024dextrah, singh2024dextrah}. Reusable latent models have recently emerged as a promising hierarchical approach for humanoid control. %
First, a low-level latent-conditioned controller (LLC) is trained to generate a diverse repertoire of motions \cite{Peng2022-vr,Tessler_undated-zi,Luo2023-er,Won2022-jy}. %
Then, a high-level controller (HLC) is trained to utilize the LLC by predicting latents, effectively reusing the learned skills. This hierarchical structure has been used in complex scenarios involving humanoid-object interactions \cite{braun2023physically, luoomnigrasp}, where the separation between high-level task planning and low-level motion execution can significantly reduce the learning complexity.

\paragraph{Lifelong Learning} Interactive agents face the challenge of unpredictable operational environments. The framework of lifelong learning addresses this by continuously acquiring and transferring knowledge across multiple tasks sequentially over the agent's lifetime \cite{silver2013lifelong}. In reinforcement learning, lifelong learning approaches have demonstrated superior performance in complex tasks through knowledge reuse and retention \cite{tessler2017deep,abel2018policy,nath2023sharing,meng2025preserving}. Our work encodes all tasks via visual cues, enabling the policy to learn new tasks through fine-tuning without architectural modifications. We show that this capability is crucial and allows the \name to master simple skills before adapting to diverse, complex scenes.

\section{Method}

To tackle the problem of visual dexterous humanoid control, we train a policy to output joint actuation based on current visual and proprioceptive observations. In~\Cref{sec:task}, we define the task settings of our visual dexterous control problem. In~\Cref{sec:vis_interface}, we describe our perception-as-interface design to enable vision-only task learning. Finally, in~\Cref{sec:learning}, we detail how we learn our control policy.

\subsection{Task Definitions}
\label{sec:task}
 We study two scenarios for visual dexterous whole-body control: (1) a tabletop setting where the humanoid manipulates objects on an isolated table floating in midair, and (2) a naturalistic kitchen environment with diverse objects, randomized placements, and articulated cabinets, as visualized in \cref{fig:kitchens}. The tabletop scenario, inspired by Omnigrasp~\cite{luoomnigrasp}, is devoid of complex visual backgrounds, providing a controlled training environment for visual-dexterous grasp learning. Here, the humanoid is initialized facing the table with one object on the table. The policy is trained to approach the table, pick up an object, move it to a designated 3D location, and place it down. The kitchen scene presents a significantly more challenging setting, featuring a photorealistic background, distractor objects, and articulated drawers. In this environment, we consider both object transport and drawer manipulation. In the object transport task, the object position is unknown, and the humanoid must first locate the target object, pick it up, transport it to the specified 3D location, and release it. In the articulated cabinets task, the agent is tasked with opening and closing specified cabinets on command.

\paragraph{Scene Generation} To create diverse environments, we leverage SceneSynthesizer \cite{Eppner2024} to generate six distinct kitchen types (galley, flat, L-shaped, with island, U-shaped, and peninsula). For each type, we produce 80 randomized configurations with variations in both structural elements (\eg, cabinet quantity and placement) and surface textures. We pre-compute 20 valid humanoid spawning positions and 64 object placement locations within each generated scene, ensuring all elements are positioned without geometric intersections or scene penetrations. We create 512 environments for training and 64 for testing. 


\begin{figure}
\begin{center}
\includegraphics[width=\linewidth]{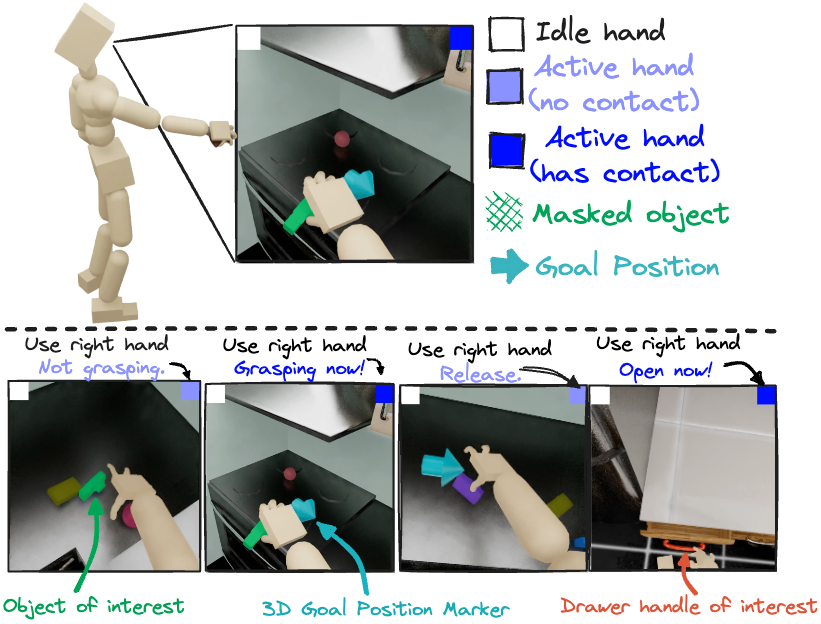}
\end{center}
\vspace{-5mm}
   \caption{\small  \textbf{Perception-as-Interface:} \name instructs the policy through visual signals. We overlay the object of interest using a green mask, use a 3D blue marker to indicate target location, and use colored 2D squares (top corners) to inform the agent which hand to use and when to grasp and release. }
    \label{fig:perception_interface}
\label{fig:arch}
\vspace{-8mm}
\end{figure}
\begin{figure*}
\begin{center}
\includegraphics[width=0.9\textwidth]{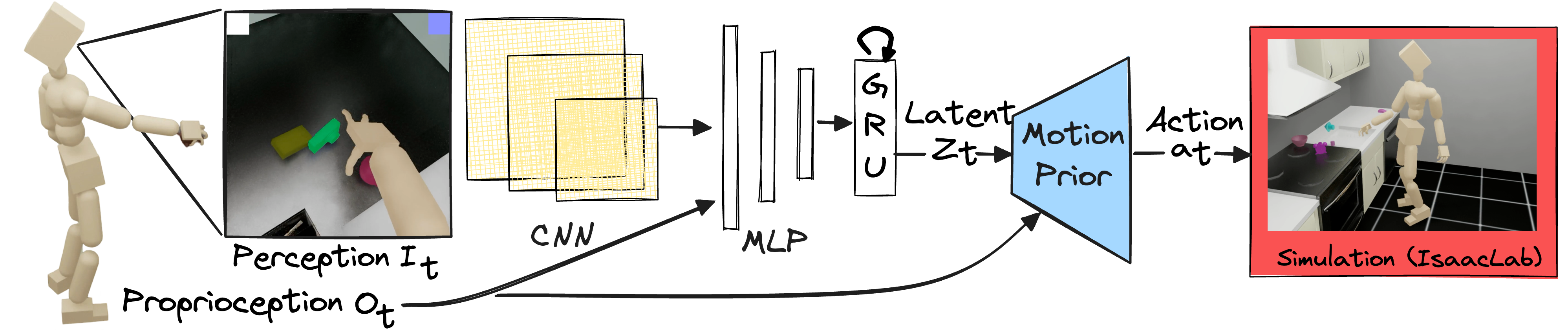}
\vspace{-2mm}
\end{center}
   \caption{\small We use perception and proprioception as input to the network, processed by a simple CNN-GRU-MLP architecture.  }
\label{fig:archi}
\vspace{-5mm}
\end{figure*}

\subsection{The Visual Perception Interface}
\label{sec:vis_interface}
To tackle the task of object search, grasp, goal reaching, release, and drawer opening, prior art resorts to phase variables \cite{Merel2020-qm} or modular state machines \cite{jiang2021dash} to decide which part of the task the policy should execute on. While adding phase variables, object information, and task instructions such as 3D goal coordinates is helpful for state-space policy training, it limits the possibility of learning new tasks.  Any change in the task requires re-training the policy to accommodate new state variables. We propose to use perception as the sole interface for task specification. Leveraging vision as a unified interface can support agents that can be adapted (through fine-tuning) to new tasks without any architectural changes. In an extreme example, visual signals could even be floating text for precise task instructions. Our approach embraces this principle, using visual indicators to specify tasks, as illustrated in \cref{fig:perception_interface}.

\paragraph{Object Selection} The kitchen scenes contain multiple objects, making object specification a significant challenge, particularly when precise location information is unavailable. For instance, selecting a particular apple from a cluster becomes ambiguous without explicit spatial coordinates. To address this challenge, we leverage 2D semantic segmentation \cite{kirillov2023segment} as a visual selection mechanism. Our approach applies alpha blending using the segmentation mask to highlight the target object with a distinctive bright green overlay directly \textbf{in the image space}. This visual differentiation enables the agent to identify the correct object and transform the object selection problem into a visually guided task. To differentiate between object grasping and drawer opening, we paint drawer handles bright red.

\paragraph{Object Reaching Goals} Once the object is picked up, the next step is specifying its destination. While 3D target locations can specify goals, such input will become redundant for tasks like object search or cabinet opening and lacks adaptability to new scenarios. To overcome these limitations, we instead \textit{render} a 3D arrow directly in the agent's visual field to indicate the desired destination. This interface resembles information markers commonly used in 3D games or AR applications and provides an intuitive and visually grounded objective for manipulation tasks.

\paragraph{Handedness, Pickup Time, and Release Time}
We train our agent for bimanual control, enabling it to use the left hand, right hand, or both hands on command. Additionally, the humanoid must be able to pick up and place objects as instructed. To achieve this, we color the visual field's upper right and left corners to indicate which hand should grasp the object. A small colored block indicates hand usage: white signals the hand should remain idle, purple designates a hand that should be used for contact but should not currently be in contact, and blue signifies active contact now (grasping). The purple signal guides the humanoid in determining which hand to use to reach out to the object. 

We believe that perception-as-interface \cite{hoffman2015interface} offers near-infinite extensibility for vision-based agents, with an extreme case being using floating text to convey instructions.

\subsection{Learning Perceptive Dexterous Control}
\label{sec:learning}
We formulate our task using the general framework of goal-conditioned RL. The learning task is formulated as a Partially Observed Markov Decision Process \cite[POMDP]{astrom1965optimal} defined by the tuple ${\mathcal M}=\langle \mathcal{\bs S}, \mathcal{\bs O}, \mathcal{ \bs A}, \mathcal{ \bs T}, \rewardfunc, \gamma\rangle$ of states, observations, actions, transition dynamics, reward function, and discount factor. The simulation determines the state $\state \in \mathcal{ \bs S}$ and transition dynamics $\mathcal{ \bs T}$, and a policy computes the action $\action \in \mathcal{ \bs A}$ based on the partial observation $\obs \in \mathcal{ \bs O}$. The observation $\obs \in \mathcal{\bs O}$ is a partial representation of the full simulation state $\state$, combined with the task instructions. We train a control policy $\policyours(\action|\obs)$ to compute joint actuation $\action$ from observations. The reward function \( r_t = \rewardfunc(\state, \action, \staten) \) is defined using the full simulation state $\state$ to guide learning. The policy optimizes the expected discounted reward $\mathbb{E}\left[\sum_{t=1}^{T} \gamma^{t-1} r_{t}\right]$ and is trained using proximal policy optimization (PPO)~\cite{Schulman2017-ft}.

\paragraph{Observation}  The policy observes the proprioception (the observation of oneself) and camera image $\obs \triangleq (\selfstate, \imaget)$. The proprioception $\selfstate \triangleq (\simr, \simt, \simlv, \simav, \contacthand)$ consists of the 3D joint rotation $\simr \in \reals^{J \times 3}$, position $\simt \in \reals^{J \times 3}$, linear velocity $\simlv \in \reals^{J \times 3}$, angular velocity $\simav \in \reals^{J \times 3}$, and the contact forces on each hand $\contacthand \in \reals^{J_H \times 3}$. Due to computational challenges posed by larger images, prior work focused on low-resolution inputs (\eg, 40$\times$30$\times$3 \cite{tirumala2024learning}); our experiments demonstrate the performance benefits of higher-resolution visual inputs. We evaluate RGB (128$\times$128$\times$3), RGBD (100$\times$100$\times$4), and stereo (2$\times$80$\times$80$\times$3), with the latter two using slightly smaller dimensions due to GPU memory constraints.

\paragraph{Reward} To train \name, we provide it with dense rewards that help and guide its behavior, as RL agents struggle when trained with sparse objectives (\eg, $r_t = \mathbf{1}_{\text{object on marker}}$). For the grasping task, we designed two rewards: a reward aimed to measure grasp and object pose, and another (optional) to guide the gaze. The grasp reward is:
\begin{equation}
\scriptsize
\label{eqn:lookat}
    \bs{r}^{\name}_t = 
    \begin{cases} 
         \rewardapproach + \rewardlookat, &   \|\reftpregrasp - \simthand\|_2 > 0.2 \;  \text{and} \; t < \contactstart \\ 
        \rewardpregrasp + \rewardlookat,  & \|\reftpregrasp - \simthand\|_2 \leq 0.2 \;  \text{and} \; t < \contactstart \\ 
        \rewardobj + \rewardlookat,  &  \contactstart  \leq t < \contactend \\
        (1- \mathbf{1}_{\text{has-contact}}),  &  t \geq \contactend   \\
    \end{cases}
\end{equation}
This reward is split into 4 sequential segments: a time scheduler (provided by the user) determines whether the agent should be in the approach, grasp, or release phase. The pre-grasp phase is broken into 2 segments, reach and pre-grasp. The pre-grasp~\cite{dasari2023learning, luoomnigrasp} is provided when the hand is $< 0.2 m$ from the target object. This encourages the hand to match a pre-computed pre-grasp $\refqpregrasp \triangleq (\reftpregrasp, \refrpregrasp)$, a single hand pose consisting of the hand translation $\reftpregrasp$ and rotation $\refrpregrasp$. If the object is successfully grasped, we switch to the object's 3D location reward $\rewardobj$ to encourage transporting the object to specific locations. After the grasping should end ($t \geq \contactend$), we encourage the agent to release the object: $\mathbf{1}_{\text{has-contact}}$ determines if any hand has contact with the object. The optional look-at reward

\begin{equation}
\small
\label{eqn:lookat}
    \rewardlookat = 
    \begin{cases} 
         \rewardlookatobject, &   \text{and} \; t < \contactstart \\ 
        \rewardlookatmarker,  & \text{and} \; t \geq \contactstart \\ 
    \end{cases} \,,
\end{equation}

is also time-conditioned. First, the humanoid is rewarded for looking at the object. After the contact start time, the humanoid is encouraged to look at the 3D target marker. The look-at reward is computed as $r_{\text{lookat}} = 1 - \sqrt{1 - (\gazevector \cdot \targetvector)}$, where $\gazevector$ is the gaze vector and $\targetvector$ is the eye to target (object or marker) vector. While we encourage the policy to always look in the object's direction, search (looking left and right) emerges when the object is not always initialized in the view. To train the drawer open policy, we use the same pregrasp and approach reward, and change the object reward position $\rewardobj$ to the drawer opening reward $\rewarddrawer$. For details about each reward term, please refer to~\Cref{sec:app_pdc}.

\paragraph{Early Termination} Equally important to reward design, early termination \cite{peng2018deepmimic} also plays an important role in shaping the agent's behavior as it provides a strong negative reward signal to the agent. We early terminate the training episode in the following scenarios: (1) the agent is not in contact with the object by $\contactstart$ grasping time, (2) the agent is still in contact of the object after contact end time $\contactend$, and (3) if the target object is more than 0.25m away from the marker for more than 2 seconds.

\begin{figure*}[t!]
    \centering
    \includegraphics[width=0.98\linewidth]{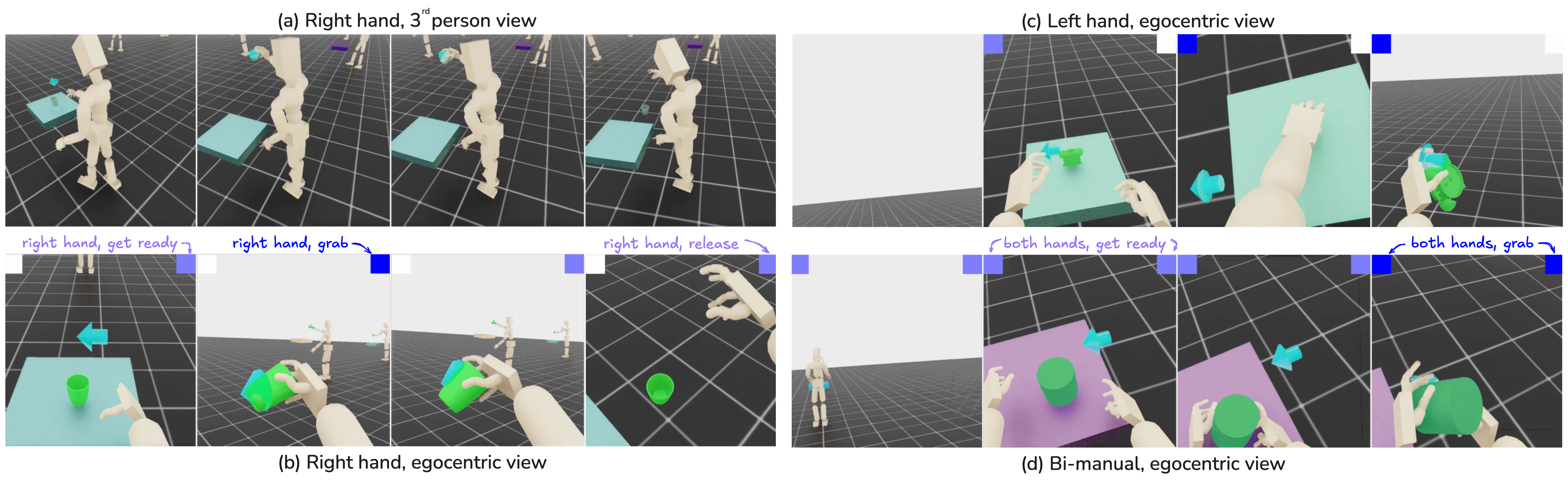}\vspace{-0.2cm}
    \caption{\small  \textbf{Tabletop:} \name is instructed directly from visual signals. A visual (top left and/or right) indicates in {\color{Periwinkle}purple} whether the corresponding hand should be prepared for contact. Changing to {\color{Blue}dark-blue} indicates that contact should be made. Instructing the agent to use both hands enables it to transport larger objects. Changing the indicator back to {\color{Periwinkle}purple} instructs the agent to release the object.}
    \label{fig: tabletop}
    \vspace{-0.2cm}
\end{figure*}

\begin{figure*}[t!]
    \centering
    \begin{subfigure}[t]{0.48\linewidth}
        \centering
        \includegraphics[width=\textwidth]{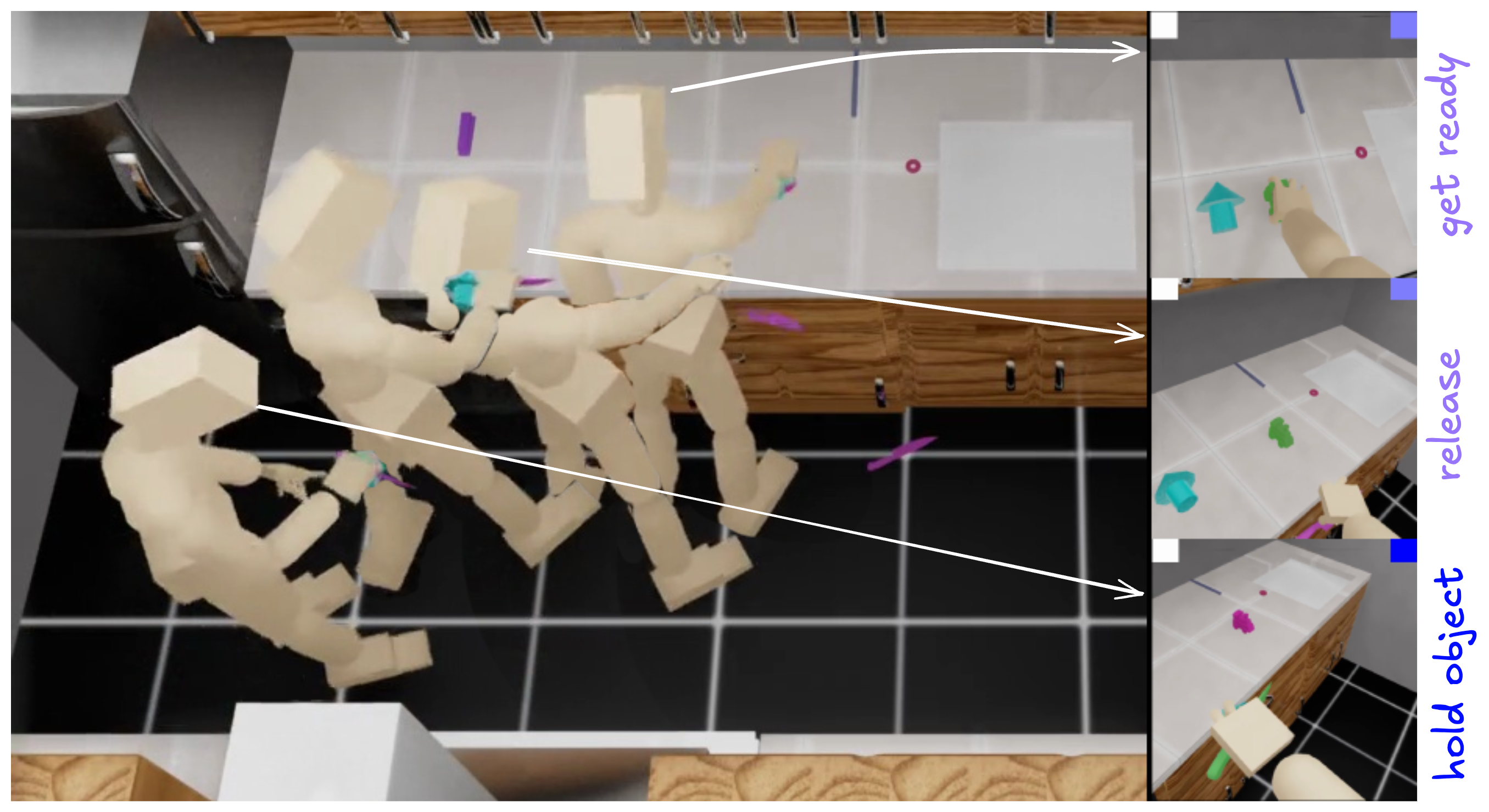}
        \caption{Multi-object manipulation.}
    \end{subfigure}%
    ~ 
    \begin{subfigure}[t]{0.46\linewidth}
        \centering
        \includegraphics[width=\textwidth]{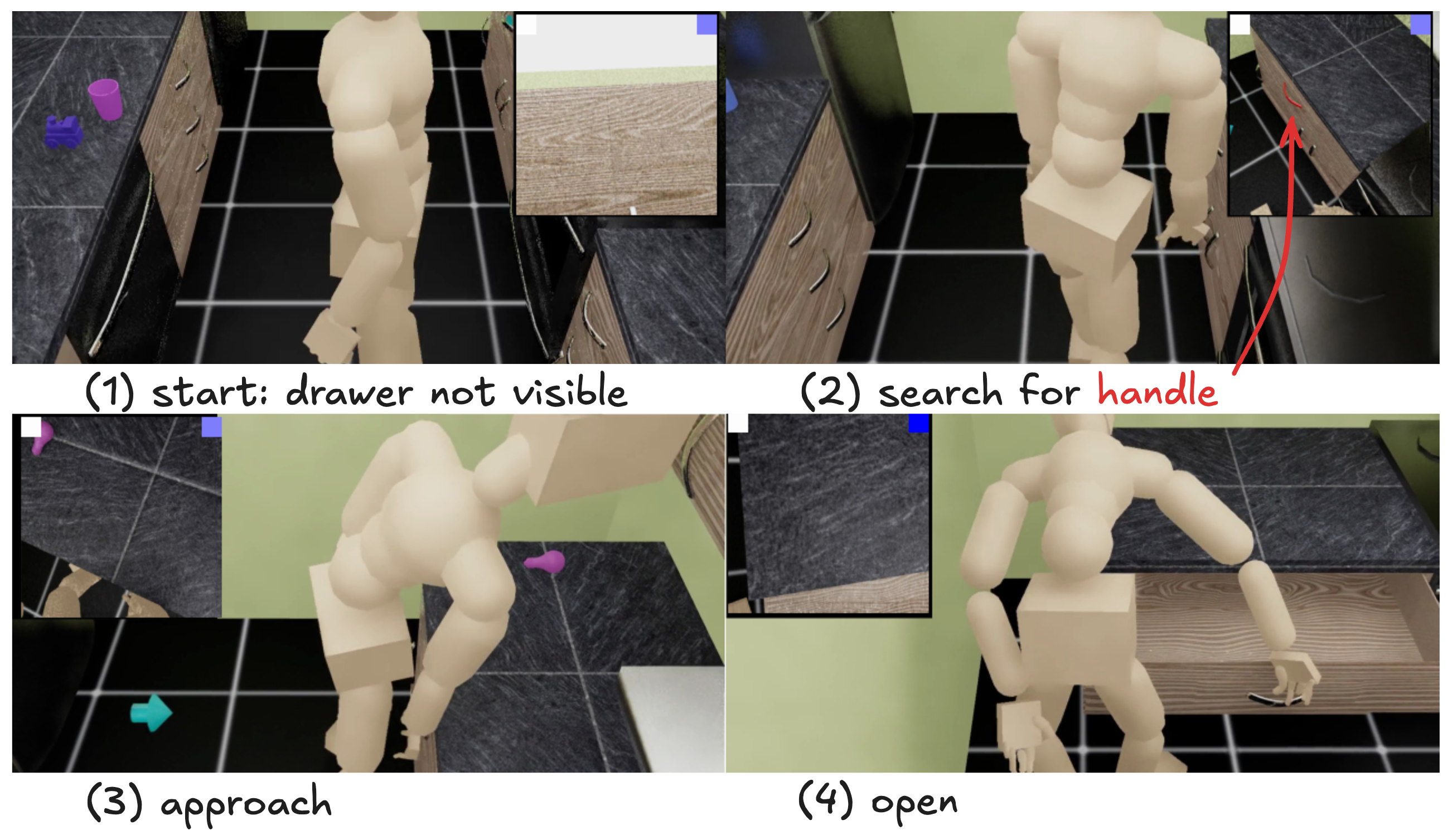}
        \caption{Search and open drawers.}
    \end{subfigure} 
    \caption{\small  \textbf{Kitchen task:} Here, the agent must master multiple skills. First, the agent must search and find the objective. A target object is marked in {\color{Green}green}, whereas a target drawer handle in {\color{Red}red}. Once found, the agent must approach the target before it can interact with it.}
    \label{fig: kitchen_task}
    \vspace{-0.4cm}
\end{figure*}

\paragraph{Policy Architecture} We use a simple CNN-GRU-MLP architecture for the policy, illustrated in \Cref{fig:archi}. The CNN processes raw image inputs, extracting high-level spatial features. The GRU provides recurrent memory, enabling the agent to handle temporal dependencies of this partially observable task. Due to the large throughput of data when training with an image-conditioned RL policy, we use a simple three-layer CNN architecture that has been used in sim-to-real robotics settings \cite{lin2025simtorealreinforcementlearningvisionbased}. For RGBD, the depth channel is treated as an additional image channel. For stereo, we use a Siamese network~\cite{koch2015siamese} and process the two images with the same CNN and concatenate the features. We also experiment with frozen pretrained visual encoders like ResNet \cite{He2015-hk} and ViT~\cite{dosovitskiy2020image} that are pretrained on classification tasks, though they perform poorly in our setting. 

\paragraph{Humanoid Motion Representation} Due to the difficulty of controlling a high-dimensional humanoid with dexterous hands, many prior works resort to reference motion imitation \cite{wang2023physhoi, xu2025intermimicuniversalwholebodycontrol} or overfitting to a single type of action \cite{bae2023pmp}. We do not use any reference motion and utilize the recently proposed PULSE-X \cite{luoomnigrasp, Luo2023-er} as the reusable, general-purpose, low-level controller. PULSE-X is a conditional VAE (cVAE) that is trained to reproduce motions from the AMASS MoCap dataset \cite{Mahmood2019-ki}. At a high level, PULSE-X learns a proprioception-conditioned latent space that can be decoded into human-like actions $\decoder(\action|\latentttask, \selfstate)$. Using PULSE-X, we define the action space of our policy with respect to the prior $\action = \decoder(\policyours(\latentttask| \selfstate, \imaget), \selfstate)$ and train it using hierarchical RL. For more information about PULSE-X, please refer to~\Cref{sec:app_pulsex}.\\

\noindent \textbf{Tabletop training:} the humanoid is initialized at a random location facing (but not looking at) the object. The target goal position is sampled randomly above the tabletop and changes periodically. \textbf{Kitchen training:} the humanoid is initialized randomly in an empty space facing a random direction. Target object positions are sampled randomly on the support surface and changed periodically (see samples in {\tt\small \href{\link}{supplement site}}). \textbf{Optimization:} Unlike imitiation-based methods~\cite{wang2023physhoi,xu2025intermimicuniversalwholebodycontrol,Won2021-sn}, we do not use any reference human motions during training. We follow the standard on-policy RL \cite[PPO]{Schulman2017-ft} training procedure, interleaving sample collection and policy updates. To successfully solve the kitchen scene, we warm-start the policy using the trained table-top policy. \textit{This is enabled} by our perception-as-interface paradigm, as we can reuse the grasping capability from the tabletop agent. For the kitchen scene, half of the environments learn the grasping task while the other half opens drawers.

\section{Experiments}
\label{sec:exp}

\begin{table*}[t]

\centering

\resizebox{\linewidth}{!}{%
\begin{tabular}{l|r|rr|rr|rr|rr|rr|rr|rr|rr}
\toprule
\multicolumn{1}{c}{} & \multicolumn{1}{c}{} &\multicolumn{8}{|c}{\cellcolor{rqblue}GRAB-Train (25 seen objects )} & \multicolumn{8}{|c}{\cellcolor{rqred}GRAB-Test (5 unseen objects))} 
\\ 
\midrule
Method  & Vision & $\success \uparrow$  &  $ \poserror \downarrow $ &  $\successright  \uparrow$ &  $ \poserror \downarrow $ &  $ \successleft \uparrow$  &  $ \poserror \downarrow $ & $\successbi  \uparrow$ &  $ \poserror \downarrow $ &  $\success \uparrow$  &  $ \poserror \downarrow $ &  $\successright  \uparrow$ &  $ \poserror \downarrow $ &  $ \successleft \uparrow$  &  $ \poserror \downarrow $ & $\successbi  \uparrow$ &  $ \poserror \downarrow $     \\ \midrule

Omnigrasp  & \xmark & {99.1\%} & {8.9} & {99.1\%} & {8.9} & {98.7\%} & {13.0} & {99.7\%} & {12.5} & {70.6\%} & {11.2} & {70.4\%} & {11.2} & {-} & {-}& {100.0\%} & {18.7} \\

\midrule 

\name-RGB  & \cmark & {87.5\%} & {74.4} & {88.7\%} & {72.6} & {83.0\%} & {77.4} & {78.7\%} & {98.3} & {90.1\%} & {100.5} & \textbf{90.0\%} & {101.0}  & - & - & \textbf{100.0\%} & {39.1}\\
\name-RGBD  & \cmark & {86.9\%} & {61.5} & {88.1\%} & {60.7} & {80.1\%} & \textbf{68.6} & {83.9\%} & \textbf{59.5} & {85.2\%} & {74.6} & {85.1\%} & {74.8} & {-} & {-} & \textbf{100.0\%} & {41.8}\\
\name-stereo & \cmark & \textbf{96.4\%} & \textbf{51.9} & \textbf{96.9\%} & \textbf{47.7} & \textbf{92.9\%} & {79.2} & \textbf{87.7\%} & {57.5} & \textbf{91.8\%} & \textbf{61.0} & {85.8\%} & \textbf{61.1} & {-} & {-} & \textbf{100.0\%} & \textbf{41.6}\\


\bottomrule 
\end{tabular}}
\caption{\small{Quantitative results on visual object grasps and target goal reaching on the GRAB dataset.  The training set contains 82.4\% right hand, 12.1\% left hand, and 5.4\% both hands. The test set contains 99\% right hand grasp and 1\% both hands. }} 
\label{tab:grab}

\end{table*}

\begin{table*}[t]

\centering
\resizebox{\linewidth}{!}{%
\begin{tabular}{l|r|r|rr|r|r|rr|r|r|rr|rr|rr}
\toprule
\multicolumn{1}{c|}{} & \multicolumn{4}{c|}{\cellcolor{rqblue}GRAB-Train, Scene-Train}  &  \multicolumn{4}{c}{\cellcolor{rqred}GRAB-Test, Scene-Train} &  \multicolumn{4}{|c}{\cellcolor{llnvgreen}GRAB-Test, Scene-Test} & \multicolumn{2}{|c}{\cellcolor{rqblue}Drawer, Scene-Train} & \multicolumn{2}{|c}{\cellcolor{llnvgreen}Drawer, Scene-Test}\\ 
\midrule
\multicolumn{1}{c|}{} & \multicolumn{1}{c|}{Search}  & \multicolumn{1}{c|}{Pick up}  & \multicolumn{2}{c|}{Goal Reaching} & \multicolumn{1}{c|}{Search}  & \multicolumn{1}{c|}{Pick up}  & \multicolumn{2}{c}{Goal Reaching} & \multicolumn{1}{|c|}{Search}  & \multicolumn{1}{c|}{Pick up}  & \multicolumn{2}{c}{Goal Reaching} & \multicolumn{1}{|c}{Search}  & \multicolumn{1}{c}{Drawer Open} & \multicolumn{1}{|c}{Search}  & \multicolumn{1}{c}{Drawer Open}
\\ 
\midrule
 $\text{Method}$     & $\successsearch \uparrow$ &   $\successgrasp \uparrow$ & $\successtraj \uparrow$ & $\poserror \uparrow$ & $\successsearch \uparrow$ &   $\successgrasp \uparrow$ & $\successtraj \uparrow$ & $\poserror \uparrow$ & $\successsearch \uparrow$ &   $\successgrasp \uparrow$ & $\successtraj \uparrow$ & $\poserror \uparrow$ & $\successsearch \uparrow$ &   $\successdrawer \uparrow$& $\successsearch \uparrow$ &   $\successdrawer \uparrow$ \\ \midrule
  \text{\name - Stereo}    & { 98.1\% } & { 85.1\% } &{ 65.5\% } & { 162.2}  & { 98.9\% } & {79.1 \% } &{ 52.8\% } & { 147.6} & {95.7\%} & {80.2} & {53.6\%} & {154.4}  & {98.8\%} & {63.7\%}  & {98.0\%} & {64.0\%}\\
 
\bottomrule 
\end{tabular}}
\caption{Quantitative results on visual object grasps, target goal reaching, and drawer opening in the kitchen scene. We test on different combination of train/test scenes and train/test objects.  }
\label{tab:kitchen}
\vspace{-5mm}
\end{table*}

\paragraph{Baselines} As few prior attempts at visual dexterous whole-body control exist, we compare against a modified version of a SOTA state-space policy, Omnigrasp~\cite{luoomnigrasp}. The state-space policy always observes the object and target location and serves as the oracle policy. We extend the original Omnigrasp policy to support hand specification and placing down the object. The state-space policy is trained using the exact same reward as the visual controllers, including the look-at reward. Please refer to \Cref{sec:app_omnigrasp} for more information on this baseline.

\paragraph{Implementation Details} Simulation is conducted in NVIDIA IsaacLab~\cite{mittal2023orbit} and rendered using tiled rendering. Due to the significantly increased data throughput for training visual RL, we simulate 192 environments per-GPU across 8 Nvidia L40 GPUs in parallel. Simulation is conducted at 120Hz while the control policy runs at 30Hz. The tabletop policy training takes around 5 days to converge. Fine-tuning a trained tabletop policy for the kitchen scene requires an additional 5 days. This results in approximately $5\times10^9$ samples, roughly 4 years' worth of experience. Our simulated humanoid follows the kinematic structure of SMPL-X~\cite{Pavlakos2019-fd} using the mean body shape \cite{Luo2021-gu, Luo2023-ft, luoomnigrasp}. This humanoid has $J = 52$ joints, of which 51 are actuated. Between these joints, 21 joints belong to the body, and the remaining $J_H = 30$ joints are for two hands. All joints are 3 degrees-of-freedom (DoF) spherical joints actuated with PD controllers, resulting in $\action \in \reals^{51 \times 3}$. To mimic human perception, we attach camera(s) to the head of the humanoid, roughly at the position of the eyes. For stereo, we place the cameras 6 cm apart, similar to human eye placements.

\paragraph{Object Setups} We use a subset of the GRAB \cite{taheri2020grab} dataset (25 out of 45 objects) that contains household objects for training and keep the testing set (5 objects) for both the kitchen and tabletop setting. We obtain the pre-grasp for the reward and hand specification from the MoCap recordings provided by the GRAB dataset: \eg if the pre-grasp uses both hands, the episode will use the corresponding hands. We also provide qualitative results on larger and more diverse objects on the OMOMO~\cite{li2023object} and Oakink~\cite{YangCVPR2022OakInk} datasets (see {\tt\small \href{\link}{supplement site}}). 

\begin{table*}

\raggedleft

\centering
\scriptsize

\resizebox{\linewidth}{!}
{%
\begin{tabular}{lrrrrrr|rr|rr|rr}
\toprule
\multicolumn{13}{c}{\cellcolor{rqblue}GRAB-Test (5 unseen objects))} 
\\ 
\midrule
 \text{idx} & $\text{Feature Extractor}$ & $\text{Resolution}$ & $\text{Input Modality}$  & $\text{Lookat Reward}$ & $\text{Distillation}$ & $\text{PULSE-X}$ &  $\successgrasp \uparrow$  &  $ \poserror \downarrow $ &  $\successright  \uparrow$ &  $ \poserror \downarrow $  & $\successbi  \uparrow$ &  $ \poserror \downarrow $  \\ \midrule

  1 & ViT & 128 & RGB &\cmark &\xmark &  \cmark & {61.4\%} & {142.9} & {61.4\%} & {142.8} & {60.0\%} &  {161.3} \\ %
  2 & ResNet & 128 & RGB& \cmark& \xmark & \cmark&   {68.5\%} & {194.7} & {68.7\%} & {194.0} & {40.0} &  {375.7} \\ %
  3 & CNN & 128 & RGB &\cmark &\cmark & \cmark&   {68.2\%} & \textbf{16.2} & {68.6} & \textbf{16.3} & {10.0\%} &  {79.0} \\ %
  4 & CNN & 128 & RGB &\cmark &\cmark & \xmark&   {71.4\%} & {161.1} & {74.1} & {41.6} & {10.0\%} &  {69.8} \\ %
  5 & CNN & 32 & RGB & \cmark&\xmark & \cmark &  {80.7\%} & {116.4} & {80.6\%} & {115.7} & \textbf{100.0\%} &  {189.9} \\ %
  6 & CNN & 64 & RGB & \cmark&\xmark & \cmark &  {80.2\%} & {100.1} & {80.1\%} &  {100.2} & {90.0\%} & {82.0}  \\ %
  7 & CNN & 128 & RGB & \xmark&\xmark & \cmark&  {89.6\%} & {85.6} & {89.5\%} & {85.9}  & {100.0\%} & \textbf{48.0}\\
  8 & CNN & 128 & RGB & \cmark&\xmark & \cmark&   \textbf{90.1\%} & {100.5} & \textbf{90.0\%} & {101.0}  & \textbf{100.0\%} & \textbf{39.1}\\
  
\bottomrule 
\end{tabular}}

\caption{\small Ablation on various strategies of training $\name$.  }\label{tab:ablation}
\vspace{-5mm}
\end{table*}

\subsection{Results}
As motion is best seen in videos, we strongly recommend that readers view the {\tt\small \href{\link}{qualitative videos}}.

\paragraph{Metrics} In the tabletop scenario, the agent is evaluated on its ability to pick up an object, track a marker located 30cm above the table, and then release the object. An episode is considered successful if the humanoid picks up the object using the correct hand, reaches the target location ($<$25cm), and then places the object back down. The sequence of actions is time-conditioned and specified through visual cues. The policy has 2 seconds to pick up the object and 2 seconds to reach the target. We report the distance of the object from the visual marker $\poserror$ (mm), after the object is picked up. We also provide a breakdown of the success rate based on which hand to use --- right $\successright$, left $\successleft$, and both hands $\successbi$. The kitchen scene presents an additional challenge. The policy must search for the object, reach and pick it up, and then follow the marker. Here, the marker moves across the counter (instead of straight lifting). We report the search performance $\successsearch$, measuring whether the object comes into view; the grasp success $\successgrasp$, measuring whether the correct object is grasped; and trajectory following success $\successtraj$, which shows whether the object follows the trajectory ($<$25cm) and is successfully released at the end. 
Finally, we measure the policy's ability to open drawers $\successdrawer$. For the kitchen setup, we test in both training scenes and \textit{unseen scenes}.

\paragraph{Tabletop: Object Lifting} We report the results in \Cref{tab:grab} and \Cref{fig: tabletop}.
The oracle state-space policy reaches a high success rate for the training set, yet its success rate drops on the test set. On the other hand, the visual policy reaches a similar success rate across both training and testing. This result shows that while the ground-truth object shape information used by the state-space policy (extracted using BPS \cite{prokudin2019efficient} using the canonical pose) can help achieve a high success rate during training, vision provides better generalization capabilities. Since most of the reference pre-grasps in the data use the right hand, \name performs better at using the right hand than using the left hand and both hands. In terms of vision format, stereo outperforms both RGBD and RGB, with RGBD coming in second. Surprisingly, stereo outperforms RGBD across all metrics, suggesting that stereo depth estimation emerges from grasping and reaching target goals.

\paragraph{Kitchen Scenes: Search, Grasp, Move, and Drawers} For the kitchen scene, as shown in \Cref{tab:kitchen} and \Cref{fig: kitchen_task}, our policy achieves a high search and grasping success rate, even when the object is deep into the counter and the humanoid is required to lean on the counter to reach the object. The goal reaching success rate is lower compared to the tabletop setting, partially due to the harder goal positions in the kitchen setting (random positions above the counter). \name is robust across unseen objects and kitchen scenes without large performance degradation on unseen objects and scenes. Finally, the drawer-opening task achieves a high success rate in finding and opening the drawer.

\section{Ablation and Analysis}
\label{sec:abla}

\paragraph{Pretrained Vision Encoders} In~\Cref{tab:ablation} Row 1 (R1), R2, and R7, we study using visual encoders frozen pretrained using ImageNet \cite{deng2009imagenet}. The results show that although the policy learns to pick up using these pretrained features, it does not reach comparable success rates to learning from scratch. We hypothesize that the features trained from ImageNet are insufficient to close the perception-action loop, and training from scratch or fine-tuning the visual feature extractor is needed to learn better visual features.

\paragraph{Distillation vs From Scratch} Comparing R3 and R8, we can see that distillation achieves a much lower success rate compared to training from scratch. Distillation does reach a better position error, indicating that predicting the visual marker position in 3D can be a good auxiliary loss.

\paragraph{Humanoid Motion Prior } R4 vs R8 shows that the humanoid motion prior (PULSE-X) can lead to better overall performance. Upon visual inspection, the behavior learned using PULSE-X is also more human-like. 

\paragraph{Resolution} Comparing R5, R6, and R8: the policy success rate increases as the resolution of the images increases. This shows that higher resolution does have benefits in terms of grasp success and goal reaching. Notice that the stereo-model using two 80x80 resolution cameras achieves the highest success rate within our computational budget. 

\paragraph{Look-at Reward} R7 vs R8 shows that without the look-at reward, the policy can still learn the table-top policy and achieves comparable results. However, the look-at reward helps shape the search behavior for the kitchen scene.

\paragraph{Analysis: Emergent Search Behavior} For the kitchen scene, our policy learns interesting emergent behavior due to the setup of the task. As can be seen from the {\tt\small \href{\link}{supplement videos}}, the policy learns to look left \& right and scans the room (sometimes turns 360) while searching for the object. Also, we observe that the humanoid learns to scan the counter to find the object, a behavior that emerges from our object spawn locations. Also, the marker is not guaranteed to be in view when the object has been grasped. We observe that the humanoid learns to {search left \& right } for the marker. Such behavior emerges from the reward to look at the object and the complex kitchen setting and is not observed in the simpler tabletop setting.

\paragraph{Analysis: Object Selection} Our masking-as-object-selection design enables \name to grasp objects of interest, even when objects of the same shape and geometry are in the view. In~\Cref{tab:kitchen}, results are reported when 6 objects are together in the scene. These results show high search and pickup rates, showing that \name can find the right object.

\paragraph{Analysis: Multi-Task Learning} Our kitchen policy is warm-started from our trained tabletop policy.  While the trained-from-scratch policy does learn to search (achieving a 64.3\% success rate in search) for the object, it does not succeed in picking up the objects. Our perception-as-interface enables continuously adapting the policy to learn in different scenarios (table top $\rightarrow$ kitchen) and different tasks (object transportation and drawer opening).

\section{Discussion}

\paragraph{Failure Cases and Limitations} The vision-centric design in \name leads to emergent search behaviors and success rates that outperform the state-based benchmark. However, the kitchen scene provides some challenging scenarios. For example, as the agent reaches towards objects located near the wall (on the far end of the counter), low-hanging cabinets and the steam-collector may interfere with the agent's vision. The lack of visual clarity may lead to failing to grasp the object. In addition, if the grasp fails, our agent does not attempt to re-grasp. Instead, it shifts its focus to tracking the marker, despite not holding the object. Our policy also tends to shake its head from time to time and could benefit from image-stabilization techniques. Future work may overcome these limitations by providing additional and looser rewards. Such as enabling re-grasping, or using more advanced vision analysis to ensure the object is in view instead of naive angle comparison.

\paragraph{Conclusions} We propose \name, a framework for learning multiple vision-guided tasks using a single policy via perception-as-interface and visual reinforcement learning. We observe that training in complex and diverse scenes leads to the emergence of behaviors such as search. As one of the first to tackle visual dexterous humanoid control, we believe our method, task setting, and perception-as-interface paradigm open many doors for future research.

{
    \small
    \bibliographystyle{ieeenat_fullname}
    \bibliography{main}

\begin{thebibliography}{73}
\providecommand{\natexlab}[1]{#1}
\providecommand{\url}[1]{\texttt{#1}}
\expandafter\ifx\csname urlstyle\endcsname\relax
  \providecommand{\doi}[1]{doi: #1}\else
  \providecommand{\doi}{doi: \begingroup \urlstyle{rm}\Url}\fi

\bibitem[Abel et~al.(2018)Abel, Jinnai, Guo, Konidaris, and Littman]{abel2018policy}
David Abel, Yuu Jinnai, Sophie~Yue Guo, George Konidaris, and Michael Littman.
\newblock Policy and value transfer in lifelong reinforcement learning.
\newblock In \emph{International conference on machine learning}, pages 20--29. PMLR, 2018.

\bibitem[Astrom(1965)]{astrom1965optimal}
Karl~J Astrom.
\newblock Optimal control of markov decision processes with incomplete state estimation.
\newblock \emph{J. Math. Anal. Applic.}, 10:\penalty0 174--205, 1965.

\bibitem[Bae et~al.(2023)Bae, Won, Lim, Min, and Kim]{bae2023pmp}
Jinseok Bae, Jungdam Won, Donggeun Lim, Cheol-Hui Min, and Young~Min Kim.
\newblock Pmp: Learning to physically interact with environments using part-wise motion priors.
\newblock In \emph{ACM SIGGRAPH 2023 Conference Proceedings}, pages 1--10, 2023.

\bibitem[Braun et~al.(2024)Braun, Christen, Kocabas, Aksan, and Hilliges]{braun2023physically}
Jona Braun, Sammy Christen, Muhammed Kocabas, Emre Aksan, and Otmar Hilliges.
\newblock Physically plausible full-body hand-object interaction synthesis.
\newblock \emph{{International Conference on 3D Vision (3DV)}}, 2024.

\bibitem[Chen et~al.(2023)Chen, Tippur, Wu, Kumar, Adelson, and Agrawal]{chen2023visual}
Tao Chen, Megha Tippur, Siyang Wu, Vikash Kumar, Edward Adelson, and Pulkit Agrawal.
\newblock Visual dexterity: In-hand reorientation of novel and complex object shapes.
\newblock \emph{Science Robotics}, 8\penalty0 (84):\penalty0 eadc9244, 2023.

\bibitem[Chen et~al.(2024)Chen, Chen, Arlaud, Laptev, and Schmid]{chen2024vividex}
Zerui Chen, Shizhe Chen, Etienne Arlaud, Ivan Laptev, and Cordelia Schmid.
\newblock Vividex: Learning vision-based dexterous manipulation from human videos.
\newblock \emph{arXiv preprint arXiv:2404.15709}, 2024.

\bibitem[Cheng et~al.(2024)Cheng, Li, Yang, Yang, and Wang]{cheng2024open}
Xuxin Cheng, Jialong Li, Shiqi Yang, Ge Yang, and Xiaolong Wang.
\newblock Open-television: Teleoperation with immersive active visual feedback.
\newblock \emph{arXiv preprint arXiv:2407.01512}, 2024.

\bibitem[Dasari et~al.(2023)Dasari, Gupta, and Kumar]{dasari2023learning}
Sudeep Dasari, Abhinav Gupta, and Vikash Kumar.
\newblock Learning dexterous manipulation from exemplar object trajectories and pre-grasps.
\newblock In \emph{2023 IEEE International Conference on Robotics and Automation (ICRA)}, pages 3889--3896. IEEE, 2023.

\bibitem[Deng et~al.(2009)Deng, Dong, Socher, Li, Li, and Fei-Fei]{deng2009imagenet}
Jia Deng, Wei Dong, Richard Socher, Li-Jia Li, Kai Li, and Li Fei-Fei.
\newblock Imagenet: A large-scale hierarchical image database.
\newblock In \emph{2009 IEEE conference on computer vision and pattern recognition}, pages 248--255. Ieee, 2009.

\bibitem[Dosovitskiy et~al.(2020)Dosovitskiy, Beyer, Kolesnikov, Weissenborn, Zhai, Unterthiner, Dehghani, Minderer, Heigold, Gelly, et~al.]{dosovitskiy2020image}
Alexey Dosovitskiy, Lucas Beyer, Alexander Kolesnikov, Dirk Weissenborn, Xiaohua Zhai, Thomas Unterthiner, Mostafa Dehghani, Matthias Minderer, Georg Heigold, Sylvain Gelly, et~al.
\newblock An image is worth 16x16 words: Transformers for image recognition at scale.
\newblock \emph{arXiv preprint arXiv:2010.11929}, 2020.

\bibitem[Elfwing et~al.(2017)Elfwing, Uchibe, and Doya]{Elfwing2017-is}
Stefan Elfwing, Eiji Uchibe, and Kenji Doya.
\newblock Sigmoid-weighted linear units for neural network function approximation in reinforcement learning.
\newblock \emph{arXiv preprint arXiv:1702.03118}, 2017.

\bibitem[Eppner et~al.(2024)Eppner, Murali, Garrett, O'Flaherty, Hermans, Yang, and Fox]{Eppner2024}
Clemens Eppner, Adithyavairavan Murali, Caelan Garrett, Rowland O'Flaherty, Tucker Hermans, Wei Yang, and Dieter Fox.
\newblock scene\_synthesizer: A python library for procedural scene generation in robot manipulation.
\newblock \emph{Journal of Open Source Software}, 2024.

\bibitem[Haarnoja et~al.(2018)Haarnoja, Hartikainen, Abbeel, and Levine]{haarnoja2018latent}
Tuomas Haarnoja, Kristian Hartikainen, Pieter Abbeel, and Sergey Levine.
\newblock Latent space policies for hierarchical reinforcement learning.
\newblock In \emph{International Conference on Machine Learning}, pages 1851--1860. PMLR, 2018.

\bibitem[Haarnoja et~al.(2023)Haarnoja, Moran, Lever, Huang, Tirumala, Wulfmeier, Humplik, Tunyasuvunakool, Siegel, Hafner, Bloesch, Hartikainen, Byravan, Hasenclever, Tassa, Sadeghi, Batchelor, Casarini, Saliceti, Game, Sreendra, Patel, Gwira, Huber, Hurley, Nori, Hadsell, and Heess]{Haarnoja2023-qz}
Tuomas Haarnoja, Ben Moran, Guy Lever, Sandy~H Huang, Dhruva Tirumala, Markus Wulfmeier, Jan Humplik, Saran Tunyasuvunakool, Noah~Y Siegel, Roland Hafner, Michael Bloesch, Kristian Hartikainen, Arunkumar Byravan, Leonard Hasenclever, Yuval Tassa, Fereshteh Sadeghi, Nathan Batchelor, Federico Casarini, Stefano Saliceti, Charles Game, Neil Sreendra, Kushal Patel, Marlon Gwira, Andrea Huber, Nicole Hurley, Francesco Nori, Raia Hadsell, and Nicolas Heess.
\newblock Learning agile soccer skills for a bipedal robot with deep reinforcement learning.
\newblock \emph{arXiv preprint arXiv:2304.13653}, 2023.

\bibitem[He et~al.(2015)He, Zhang, Ren, and Sun]{He2015-hk}
Kaiming He, Xiangyu Zhang, Shaoqing Ren, and Jian Sun.
\newblock Deep residual learning for image recognition.
\newblock \emph{arXiv preprint arXiv:1512.03385}, 2015.

\bibitem[He et~al.(2024)He, Luo, He, Xiao, Zhang, Zhang, Kitani, Liu, and Shi]{he2024omnih2o}
Tairan He, Zhengyi Luo, Xialin He, Wenli Xiao, Chong Zhang, Weinan Zhang, Kris Kitani, Changliu Liu, and Guanya Shi.
\newblock Omnih2o: Universal and dexterous human-to-humanoid whole-body teleoperation and learning.
\newblock In \emph{arXiv}, 2024.

\bibitem[Hoffman et~al.(2015)Hoffman, Singh, and Prakash]{hoffman2015interface}
Donald~D Hoffman, Manish Singh, and Chetan Prakash.
\newblock The interface theory of perception.
\newblock \emph{Psychonomic bulletin \& review}, 22:\penalty0 1480--1506, 2015.

\bibitem[Huang et~al.(2024)Huang, Wang, Li, Zhang, and Fei-Fei]{huang2024rekep}
Wenlong Huang, Chen Wang, Yunzhu Li, Ruohan Zhang, and Li Fei-Fei.
\newblock Rekep: Spatio-temporal reasoning of relational keypoint constraints for robotic manipulation.
\newblock \emph{arXiv preprint arXiv:2409.01652}, 2024.

\bibitem[Jiang et~al.(2021)Jiang, Guo, Li, Exarchos, Wu, and Liu]{jiang2021dash}
Yifeng Jiang, Michelle Guo, Jiangshan Li, Ioannis Exarchos, Jiajun Wu, and C~Karen Liu.
\newblock Dash: Modularized human manipulation simulation with vision and language for embodied ai.
\newblock In \emph{Proceedings of the ACM SIGGRAPH/Eurographics Symposium on Computer Animation}, pages 1--12, 2021.

\bibitem[Kingma and Welling(2014)]{Kingma2014-vh}
Diederik~P Kingma and Max Welling.
\newblock Auto-encoding variational bayes.
\newblock \emph{2nd International Conference on Learning Representations, ICLR 2014 - Conference Track Proceedings}, pages 1--14, 2014.

\bibitem[Kirillov et~al.(2023)Kirillov, Mintun, Ravi, Mao, Rolland, Gustafson, Xiao, Whitehead, Berg, Lo, et~al.]{kirillov2023segment}
Alexander Kirillov, Eric Mintun, Nikhila Ravi, Hanzi Mao, Chloe Rolland, Laura Gustafson, Tete Xiao, Spencer Whitehead, Alexander~C Berg, Wan-Yen Lo, et~al.
\newblock Segment anything.
\newblock In \emph{Proceedings of the IEEE/CVF international conference on computer vision}, pages 4015--4026, 2023.

\bibitem[Koch et~al.(2015)Koch, Zemel, Salakhutdinov, et~al.]{koch2015siamese}
Gregory Koch, Richard Zemel, Ruslan Salakhutdinov, et~al.
\newblock Siamese neural networks for one-shot image recognition.
\newblock In \emph{ICML deep learning workshop}, pages 1--30. Lille, 2015.

\bibitem[Li et~al.(2023)Li, Wu, and Liu]{li2023object}
Jiaman Li, Jiajun Wu, and C~Karen Liu.
\newblock Object motion guided human motion synthesis.
\newblock \emph{ACM Transactions on Graphics (TOG)}, 42\penalty0 (6):\penalty0 1--11, 2023.

\bibitem[Lin et~al.(2024{\natexlab{a}})Lin, Yin, Qi, Abbeel, and Malik]{lin2024twisting}
Toru Lin, Zhao-Heng Yin, Haozhi Qi, Pieter Abbeel, and Jitendra Malik.
\newblock Twisting lids off with two hands.
\newblock \emph{arXiv:2403.02338}, 2024{\natexlab{a}}.

\bibitem[Lin et~al.(2024{\natexlab{b}})Lin, Zhang, Li, Qi, Yi, Levine, and Malik]{lin2024learning}
Toru Lin, Yu Zhang, Qiyang Li, Haozhi Qi, Brent Yi, Sergey Levine, and Jitendra Malik.
\newblock Learning visuotactile skills with two multifingered hands.
\newblock \emph{arXiv preprint arXiv:2404.16823}, 2024{\natexlab{b}}.

\bibitem[Lin et~al.(2025{\natexlab{a}})Lin, Sachdev, Fan, Malik, and Zhu]{lin2025sim}
Toru Lin, Kartik Sachdev, Linxi Fan, Jitendra Malik, and Yuke Zhu.
\newblock Sim-to-real reinforcement learning for vision-based dexterous manipulation on humanoids.
\newblock \emph{arXiv preprint arXiv:2502.20396}, 2025{\natexlab{a}}.

\bibitem[Lin et~al.(2025{\natexlab{b}})Lin, Sachdev, Fan, Malik, and Zhu]{lin2025simtorealreinforcementlearningvisionbased}
Toru Lin, Kartik Sachdev, Linxi Fan, Jitendra Malik, and Yuke Zhu.
\newblock Sim-to-real reinforcement learning for vision-based dexterous manipulation on humanoids, 2025{\natexlab{b}}.

\bibitem[Lu et~al.(2024)Lu, Cheng, Li, Yang, Ji, Yuan, Yang, Yi, and Wang]{lu2024mobile}
Chenhao Lu, Xuxin Cheng, Jialong Li, Shiqi Yang, Mazeyu Ji, Chengjing Yuan, Ge Yang, Sha Yi, and Xiaolong Wang.
\newblock Mobile-television: Predictive motion priors for humanoid whole-body control.
\newblock \emph{arXiv preprint arXiv:2412.07773}, 2024.

\bibitem[Lum et~al.(2024)Lum, Matak, Makoviychuk, Handa, Allshire, Hermans, Ratliff, and Van~Wyk]{lum2024dextrah}
Tyler Ga~Wei Lum, Martin Matak, Viktor Makoviychuk, Ankur Handa, Arthur Allshire, Tucker Hermans, Nathan~D Ratliff, and Karl Van~Wyk.
\newblock Dextrah-g: Pixels-to-action dexterous arm-hand grasping with geometric fabrics.
\newblock \emph{arXiv preprint arXiv:2407.02274}, 2024.

\bibitem[Luo et~al.()Luo, Cao, Christen, Winkler, Kitani, and Xu]{luoomnigrasp}
Zhengyi Luo, Jinkun Cao, Sammy Christen, Alexander Winkler, Kris~M Kitani, and Weipeng Xu.
\newblock Omnigrasp: Grasping diverse objects with simulated humanoids.
\newblock In \emph{The Thirty-eighth Annual Conference on Neural Information Processing Systems}.

\bibitem[Luo et~al.(2021)Luo, Hachiuma, Yuan, and Kitani]{Luo2021-gu}
Zhengyi Luo, Ryo Hachiuma, Ye Yuan, and Kris Kitani.
\newblock Dynamics-regulated kinematic policy for egocentric pose estimation.
\newblock \emph{NeurIPS}, 34:\penalty0 25019--25032, 2021.

\bibitem[Luo et~al.(2023{\natexlab{a}})Luo, Cao, Merel, Winkler, Huang, Kitani, and Xu]{Luo2023-er}
Zhengyi Luo, Jinkun Cao, Josh Merel, Alexander Winkler, Jing Huang, Kris Kitani, and Weipeng Xu.
\newblock Universal humanoid motion representations for physics-based control.
\newblock \emph{arXiv preprint arXiv:2310.04582}, 2023{\natexlab{a}}.

\bibitem[Luo et~al.(2023{\natexlab{b}})Luo, Cao, Winkler, Kitani, and Xu]{Luo2023-ft}
Zhengyi Luo, Jinkun Cao, Alexander~W. Winkler, Kris Kitani, and Weipeng Xu.
\newblock Perpetual humanoid control for real-time simulated avatars.
\newblock In \emph{International Conference on Computer Vision (ICCV)}, 2023{\natexlab{b}}.

\bibitem[Mahmood et~al.(2019)Mahmood, Ghorbani, Troje, Pons-Moll, and Black]{Mahmood2019-ki}
Naureen Mahmood, Nima Ghorbani, Nikolaus~F Troje, Gerard Pons-Moll, and Michael~J Black.
\newblock Amass: Archive of motion capture as surface shapes.
\newblock \emph{Proceedings of the IEEE International Conference on Computer Vision}, 2019-Octob:\penalty0 5441--5450, 2019.

\bibitem[Makoviychuk et~al.(2021)Makoviychuk, Wawrzyniak, Guo, Lu, Storey, Macklin, Hoeller, Rudin, Allshire, Handa, and {Gavriel State}]{Makoviychuk2021-sx}
Viktor Makoviychuk, Lukasz Wawrzyniak, Yunrong Guo, Michelle Lu, Kier Storey, Miles Macklin, David Hoeller, Nikita Rudin, Arthur Allshire, Ankur Handa, and {Gavriel State}.
\newblock Isaac gym: High performance gpu-based physics simulation for robot learning.
\newblock \emph{arXiv preprint arXiv:2108.10470}, 2021.

\bibitem[Meng et~al.(2025)Meng, Bing, Yao, Chen, Huang, Gao, Sun, and Knoll]{meng2025preserving}
Yuan Meng, Zhenshan Bing, Xiangtong Yao, Kejia Chen, Kai Huang, Yang Gao, Fuchun Sun, and Alois Knoll.
\newblock Preserving and combining knowledge in robotic lifelong reinforcement learning.
\newblock \emph{Nature Machine Intelligence}, pages 1--14, 2025.

\bibitem[Merel et~al.(2018)Merel, Ahuja, Pham, Tunyasuvunakool, Liu, Tirumala, Heess, and Wayne]{Merel2018-ah}
Josh Merel, Arun Ahuja, Vu Pham, Saran Tunyasuvunakool, Siqi Liu, Dhruva Tirumala, Nicolas Heess, and Greg Wayne.
\newblock Hierarchical visuomotor control of humanoids.
\newblock \emph{arXiv preprint arXiv:1811.09656}, 2018.

\bibitem[Merel et~al.(2020)Merel, Tunyasuvunakool, Ahuja, Tassa, Hasenclever, Pham, Erez, Wayne, and Heess]{Merel2020-qm}
Josh Merel, Saran Tunyasuvunakool, Arun Ahuja, Yuval Tassa, Leonard Hasenclever, Vu Pham, Tom Erez, Greg Wayne, and Nicolas Heess.
\newblock Catch and carry: Reusable neural controllers for vision-guided whole-body tasks.
\newblock \emph{ACM Trans. Graph.}, 39, 2020.

\bibitem[Mittal et~al.(2023)Mittal, Yu, Yu, Liu, Rudin, Hoeller, Yuan, Singh, Guo, Mazhar, Mandlekar, Babich, State, Hutter, and Garg]{mittal2023orbit}
Mayank Mittal, Calvin Yu, Qinxi Yu, Jingzhou Liu, Nikita Rudin, David Hoeller, Jia~Lin Yuan, Ritvik Singh, Yunrong Guo, Hammad Mazhar, Ajay Mandlekar, Buck Babich, Gavriel State, Marco Hutter, and Animesh Garg.
\newblock Orbit: A unified simulation framework for interactive robot learning environments.
\newblock \emph{IEEE Robotics and Automation Letters}, 8\penalty0 (6):\penalty0 3740--3747, 2023.

\bibitem[Moon et~al.(2023)Moon, Saito, Xu, Joshi, Buffalini, Bellan, Rosen, Richardson, Mallorie, Bree, Simon, Peng, Garg, McPhail, and Shiratori]{moon2023reinterhand}
Gyeongsik Moon, Shunsuke Saito, Weipeng Xu, Rohan Joshi, Julia Buffalini, Harley Bellan, Nicholas Rosen, Jesse Richardson, Mize Mallorie, Philippe Bree, Tomas Simon, Bo Peng, Shubham Garg, Kevyn McPhail, and Takaaki Shiratori.
\newblock A dataset of relighted {3D} interacting hands.
\newblock In \emph{NeurIPS Track on Datasets and Benchmarks}, 2023.

\bibitem[Nasiriany et~al.(2024)Nasiriany, Xia, Yu, Xiao, Liang, Dasgupta, Xie, Driess, Wahid, Xu, et~al.]{nasiriany2024pivot}
Soroush Nasiriany, Fei Xia, Wenhao Yu, Ted Xiao, Jacky Liang, Ishita Dasgupta, Annie Xie, Danny Driess, Ayzaan Wahid, Zhuo Xu, et~al.
\newblock Pivot: Iterative visual prompting elicits actionable knowledge for vlms.
\newblock \emph{arXiv preprint arXiv:2402.07872}, 2024.

\bibitem[Nath et~al.(2023)Nath, Peridis, Ben-Iwhiwhu, Liu, Dora, Liu, Kolouri, and Soltoggio]{nath2023sharing}
Saptarshi Nath, Christos Peridis, Eseoghene Ben-Iwhiwhu, Xinran Liu, Shirin Dora, Cong Liu, Soheil Kolouri, and Andrea Soltoggio.
\newblock Sharing lifelong reinforcement learning knowledge via modulating masks.
\newblock In \emph{Conference on Lifelong Learning Agents}, pages 936--960. PMLR, 2023.

\bibitem[No{\"e}(2004)]{noe2004action}
Alva No{\"e}.
\newblock \emph{Action in perception}.
\newblock MIT press, 2004.

\bibitem[O'regan and No{\"e}(2001)]{o2001sensorimotor}
J~Kevin O'regan and Alva No{\"e}.
\newblock A sensorimotor account of vision and visual consciousness.
\newblock \emph{Behavioral and brain sciences}, 24\penalty0 (5):\penalty0 939--973, 2001.

\bibitem[Pavlakos et~al.(2019)Pavlakos, Choutas, Ghorbani, Bolkart, Osman, Tzionas, and Black]{Pavlakos2019-fd}
Georgios Pavlakos, Vasileios Choutas, Nima Ghorbani, Timo Bolkart, Ahmed A~A Osman, Dimitrios Tzionas, and Michael~J Black.
\newblock Expressive body capture: 3d hands, face, and body from a single image.
\newblock \emph{Proceedings of the IEEE Computer Society Conference on Computer Vision and Pattern Recognition}, 2019-June:\penalty0 10967--10977, 2019.

\bibitem[Peng et~al.(2017)Peng, Berseth, Yin, and Van De~Panne]{Peng2017-il}
Xue~Bin Peng, Glen Berseth, Kangkang Yin, and Michiel Van De~Panne.
\newblock Deeploco: dynamic locomotion skills using hierarchical deep reinforcement learning.
\newblock \emph{ACM Trans. Graph.}, 36:\penalty0 1--13, 2017.

\bibitem[Peng et~al.(2018)Peng, Abbeel, Levine, and Van~de Panne]{peng2018deepmimic}
Xue~Bin Peng, Pieter Abbeel, Sergey Levine, and Michiel Van~de Panne.
\newblock Deepmimic: Example-guided deep reinforcement learning of physics-based character skills.
\newblock \emph{ACM Transactions On Graphics (TOG)}, 37\penalty0 (4):\penalty0 1--14, 2018.

\bibitem[Peng et~al.(2022)Peng, Guo, Halper, Levine, and Fidler]{Peng2022-vr}
Xue~Bin Peng, Yunrong Guo, Lina Halper, Sergey Levine, and Sanja Fidler.
\newblock Ase: Large-scale reusable adversarial skill embeddings for physically simulated characters.
\newblock \emph{arXiv preprint arXiv:2205.01906}, 2022.

\bibitem[Prokudin et~al.(2019)Prokudin, Lassner, and Romero]{prokudin2019efficient}
Sergey Prokudin, Christoph Lassner, and Javier Romero.
\newblock Efficient learning on point clouds with basis point sets.
\newblock In \emph{Proceedings of the IEEE/CVF international conference on computer vision}, pages 4332--4341, 2019.

\bibitem[Qi et~al.(2023)Qi, Yi, Suresh, Lambeta, Ma, Calandra, and Malik]{qi2023general}
Haozhi Qi, Brent Yi, Sudharshan Suresh, Mike Lambeta, Yi Ma, Roberto Calandra, and Jitendra Malik.
\newblock General in-hand object rotation with vision and touch.
\newblock In \emph{Conference on Robot Learning}, pages 2549--2564. PMLR, 2023.

\bibitem[Qin et~al.(2022)Qin, Wu, Liu, Jiang, Yang, Fu, and Wang]{qin2022dexmv}
Yuzhe Qin, Yueh-Hua Wu, Shaowei Liu, Hanwen Jiang, Ruihan Yang, Yang Fu, and Xiaolong Wang.
\newblock Dexmv: Imitation learning for dexterous manipulation from human videos.
\newblock In \emph{European Conference on Computer Vision}, pages 570--587. Springer, 2022.

\bibitem[Qin et~al.(2023)Qin, Yang, Huang, Van~Wyk, Su, Wang, Chao, and Fox]{qin2023anyteleop}
Yuzhe Qin, Wei Yang, Binghao Huang, Karl Van~Wyk, Hao Su, Xiaolong Wang, Yu-Wei Chao, and Dieter Fox.
\newblock Anyteleop: A general vision-based dexterous robot arm-hand teleoperation system.
\newblock \emph{arXiv preprint arXiv:2307.04577}, 2023.

\bibitem[Ross et~al.(2010)Ross, Gordon, and Bagnell]{Ross2010-cc}
Stephane Ross, Geoffrey~J Gordon, and J~Andrew Bagnell.
\newblock A reduction of imitation learning and structured prediction to no-regret online learning.
\newblock \emph{arXiv preprint arXiv:1011.0686}, 2010.

\bibitem[Schulman et~al.(2017)Schulman, Wolski, Dhariwal, Radford, and Klimov]{Schulman2017-ft}
John Schulman, Filip Wolski, Prafulla Dhariwal, Alec Radford, and Oleg Klimov.
\newblock Proximal policy optimization algorithms, 2017.

\bibitem[Silver et~al.(2013)Silver, Yang, and Li]{silver2013lifelong}
Daniel~L Silver, Qiang Yang, and Lianghao Li.
\newblock Lifelong machine learning systems: Beyond learning algorithms.
\newblock In \emph{AAAI Spring Symposium: Lifelong Machine Learning}, 2013.

\bibitem[Singh et~al.(2024{\natexlab{a}})Singh, Loquercio, Sferrazza, Wu, Qi, Abbeel, and Malik]{singh2024hand}
Himanshu~Gaurav Singh, Antonio Loquercio, Carmelo Sferrazza, Jane Wu, Haozhi Qi, Pieter Abbeel, and Jitendra Malik.
\newblock Hand-object interaction pretraining from videos.
\newblock \emph{arXiv preprint arXiv:2409.08273}, 2024{\natexlab{a}}.

\bibitem[Singh et~al.(2024{\natexlab{b}})Singh, Allshire, Handa, Ratliff, and Van~Wyk]{singh2024dextrah}
Ritvik Singh, Arthur Allshire, Ankur Handa, Nathan Ratliff, and Karl Van~Wyk.
\newblock Dextrah-rgb: Visuomotor policies to grasp anything with dexterous hands.
\newblock \emph{arXiv preprint arXiv:2412.01791}, 2024{\natexlab{b}}.

\bibitem[Sutton et~al.(1999)Sutton, Precup, and Singh]{sutton1999between}
Richard~S Sutton, Doina Precup, and Satinder Singh.
\newblock Between mdps and semi-mdps: A framework for temporal abstraction in reinforcement learning.
\newblock \emph{Artificial intelligence}, 112\penalty0 (1-2):\penalty0 181--211, 1999.

\bibitem[Taheri et~al.(2020)Taheri, Ghorbani, Black, and Tzionas]{taheri2020grab}
Omid Taheri, Nima Ghorbani, Michael~J Black, and Dimitrios Tzionas.
\newblock Grab: A dataset of whole-body human grasping of objects.
\newblock In \emph{Computer Vision--ECCV 2020: 16th European Conference, Glasgow, UK, August 23--28, 2020, Proceedings, Part IV 16}, pages 581--600. Springer, 2020.

\bibitem[Tessler et~al.()Tessler, Yoni~Kasten, Guo, and Nvidia]{Tessler_undated-zi}
Chen Tessler, Israel Yoni~Kasten, Israel~Yunrong Guo, and Canada Nvidia.
\newblock Calm: Conditional adversarial latent models for directable virtual characters.

\bibitem[Tessler et~al.(2017)Tessler, Givony, Zahavy, Mankowitz, and Mannor]{tessler2017deep}
Chen Tessler, Shahar Givony, Tom Zahavy, Daniel Mankowitz, and Shie Mannor.
\newblock A deep hierarchical approach to lifelong learning in minecraft.
\newblock In \emph{Proceedings of the AAAI conference on artificial intelligence}, 2017.

\bibitem[Tirumala et~al.(2024)Tirumala, Wulfmeier, Moran, Huang, Humplik, Lever, Haarnoja, Hasenclever, Byravan, Batchelor, et~al.]{tirumala2024learning}
Dhruva Tirumala, Markus Wulfmeier, Ben Moran, Sandy Huang, Jan Humplik, Guy Lever, Tuomas Haarnoja, Leonard Hasenclever, Arunkumar Byravan, Nathan Batchelor, et~al.
\newblock Learning robot soccer from egocentric vision with deep reinforcement learning.
\newblock \emph{arXiv preprint arXiv:2405.02425}, 2024.

\bibitem[Wan et~al.(2023)Wan, Geng, Liu, Shan, Yang, Yi, and Wang]{wan2023unidexgrasp++}
Weikang Wan, Haoran Geng, Yun Liu, Zikang Shan, Yaodong Yang, Li Yi, and He Wang.
\newblock Unidexgrasp++: Improving dexterous grasping policy learning via geometry-aware curriculum and iterative generalist-specialist learning.
\newblock In \emph{Proceedings of the IEEE/CVF International Conference on Computer Vision}, pages 3891--3902, 2023.

\bibitem[Wang et~al.(2023)Wang, Lin, Zeng, Luo, Zhang, and Zhang]{wang2023physhoi}
Yinhuai Wang, Jing Lin, Ailing Zeng, Zhengyi Luo, Jian Zhang, and Lei Zhang.
\newblock Physhoi: Physics-based imitation of dynamic human-object interaction.
\newblock \emph{arXiv preprint arXiv:2312.04393}, 2023.

\bibitem[Won et~al.(2021)Won, Gopinath, and Hodgins]{Won2021-sn}
Jungdam Won, Deepak Gopinath, and Jessica Hodgins.
\newblock Control strategies for physically simulated characters performing two-player competitive sports.
\newblock \emph{ACM Trans. Graph.}, 40:\penalty0 1--11, 2021.

\bibitem[Won et~al.(2022)Won, Gopinath, and Hodgins]{Won2022-jy}
Jungdam Won, Deepak Gopinath, and Jessica Hodgins.
\newblock Physics-based character controllers using conditional vaes.
\newblock \emph{ACM Trans. Graph.}, 41:\penalty0 1--12, 2022.

\bibitem[Xie et~al.(2023)Xie, Tseng, Starke, van~de Panne, and Liu]{xie2023hierarchical}
Zhaoming Xie, Jonathan Tseng, Sebastian Starke, Michiel van~de Panne, and C~Karen Liu.
\newblock Hierarchical planning and control for box loco-manipulation.
\newblock \emph{Proceedings of the ACM on Computer Graphics and Interactive Techniques}, 6\penalty0 (3):\penalty0 1--18, 2023.

\bibitem[Xu et~al.(2023{\natexlab{a}})Xu, Hu, Doshi, Rovinsky, Kumar, Gupta, and Levine]{xu2023dexterous}
Kelvin Xu, Zheyuan Hu, Ria Doshi, Aaron Rovinsky, Vikash Kumar, Abhishek Gupta, and Sergey Levine.
\newblock Dexterous manipulation from images: Autonomous real-world rl via substep guidance.
\newblock In \emph{2023 IEEE International Conference on Robotics and Automation (ICRA)}, pages 5938--5945. IEEE, 2023{\natexlab{a}}.

\bibitem[Xu et~al.(2024)Xu, Xu, Xu, Chi, Wetzstein, Veloso, and Song]{xu2024flow}
Mengda Xu, Zhenjia Xu, Yinghao Xu, Cheng Chi, Gordon Wetzstein, Manuela Veloso, and Shuran Song.
\newblock Flow as the cross-domain manipulation interface.
\newblock \emph{arXiv preprint arXiv:2407.15208}, 2024.

\bibitem[Xu et~al.(2025{\natexlab{a}})Xu, Ling, Wang, and Gui]{xu2025intermimicuniversalwholebodycontrol}
Sirui Xu, Hung~Yu Ling, Yu-Xiong Wang, and Liang-Yan Gui.
\newblock Intermimic: Towards universal whole-body control for physics-based human-object interactions, 2025{\natexlab{a}}.

\bibitem[Xu et~al.(2025{\natexlab{b}})Xu, Zhang, Li, Han, and Lu]{xu2025humanvla}
Xinyu Xu, Yizheng Zhang, Yong-Lu Li, Lei Han, and Cewu Lu.
\newblock Humanvla: Towards vision-language directed object rearrangement by physical humanoid.
\newblock \emph{Advances in Neural Information Processing Systems}, 37:\penalty0 18633--18659, 2025{\natexlab{b}}.

\bibitem[Xu et~al.(2023{\natexlab{b}})Xu, Wan, Zhang, Liu, Shan, Shen, Wang, Geng, Weng, Chen, et~al.]{xu2023unidexgrasp}
Yinzhen Xu, Weikang Wan, Jialiang Zhang, Haoran Liu, Zikang Shan, Hao Shen, Ruicheng Wang, Haoran Geng, Yijia Weng, Jiayi Chen, et~al.
\newblock Unidexgrasp: Universal robotic dexterous grasping via learning diverse proposal generation and goal-conditioned policy.
\newblock In \emph{Proceedings of the IEEE/CVF Conference on Computer Vision and Pattern Recognition}, pages 4737--4746, 2023{\natexlab{b}}.

\bibitem[Yang et~al.(2022)Yang, Li, Zhan, Wu, Xu, Liu, and Lu]{YangCVPR2022OakInk}
Lixin Yang, Kailin Li, Xinyu Zhan, Fei Wu, Anran Xu, Liu Liu, and Cewu Lu.
\newblock {OakInk}: A large-scale knowledge repository for understanding hand-object interaction.
\newblock In \emph{IEEE/CVF Conference on Computer Vision and Pattern Recognition (CVPR)}, 2022.

\end{thebibliography}
}

\newpage
\clearpage
\appendix{   
    \hypersetup{linkcolor=black}
    \begin{Large}
        \textbf{Appendix}
    \end{Large}
    \etocdepthtag.toc{mtappendix}
    \etocsettagdepth{mtchapter}{none}
    \etocsettagdepth{mtappendix}{subsection}
    \newlength\tocrulewidth
    \setlength{\tocrulewidth}{1.5pt}
    \parindent=0em
    \etocsettocstyle{\vskip0.5\baselineskip}{}
    \tableofcontents
}

\section*{Appendices}

\section{Introduction}
In this document, we include additional details about \name that are omitted from the main paper due to space limits. In~\Cref{sec:app_site}, we provide descriptions of the results shown in our {\tt\small \href{\link}{supplement site}}. In~\Cref{sec:app_hyper}, we include implementation details about the state-space oracle policy (\Cref{sec:app_omnigrasp}) and humanoid motion prior (\Cref{sec:app_pulsex}). In~\Cref{sec:app_res}, we include additional results like per-object success rate breakdown.  The code and models will be open-sourced. 

\section{Supplementary Website}
\label{sec:app_site}

As the resulting behaviors are hard to convey through text and images, we also provide {\tt\small \href{\link}{supplement videos}}.

\subsection{Teaser}
The teaser video on the top illustrates the diverse environments in which \name is trained. Across 6 different types of kitchens, spanning 80 random configurations and textures each. This diversity enables our agent to learn general behaviors, resulting in emergent search and dexterity.

\subsection{Tabletop}
The tabletop videos showcase the 3 control schemes. We provide a 3rd person video in addition to the agent's own visual (rendered at a higher quality for ease of viewing). The visual markers in the top corners of the agent's view indicate what each hand should do --- stay idle, be ready to engage, engage, and release. We show examples of right hand, left hand, and also both-hand manipulation.

This is not limited to seen objects. Our \name agent generalizes to new and unseen objects, directly from vision.

\subsection{Kitchens}

\subsubsection{Object transport}
In the kitchen demonstrations, we first showcase the task of object transport.
Here, the agent needs to identify the target object. There are 5 distractors (objects that are not the target). The target object is marked using a green mask overlay.

Once the object is picked up, the agent needs to find the target marker and bring the object towards it.

Unlike the tabletop which mainly focuses on lifting the object, here the agent needs to transport the object to a new location in the kitchen. Every 300 frames (10 seconds) we instruct it to release the object and randomly sample a new target object.

These scenes were not observed during training, showcasing the abilities of our agent to generalize. It exhibits traits such as search, where it scans the scene for the object. Moreover, an interesting emergent property is that it scans the counter-tops, as it has learned that is where kitchen objects are located (in our tasks).

\subsubsection{Articulated drawers}

Following the object transport, we showcase the ability of the agent to open drawers. These are highlighted in the agents point of view using a red overlay. The agent learns to insert its fingers through the handle in order to obtain a better grip. It then leans back with its body in order to pull the drawer open.

\subsection{Failure cases}
Finally we showcase some common failure cases. For example, although our interface enables selecting which hand to utilize, it requires a system to determine the hand-to-use apriori. When an object can not be picked up with the selected hand, the agent will fail. In these videos we show what happens when a single hand is selected to pick up a large object. The same objects are successfully picked up in the tabletop examples above, using both hands, yet fails when attempted with a single hand.

\section{Implementation Details}
\label{sec:app_hyper}
\begin{table*}[t]
\centering

\resizebox{\linewidth}{!}{%
\begin{tabular}{lcccccccccc}
\toprule
\multicolumn{11}{c}{\cellcolor{rqblue}{GRAB-Train (25 Seen Objects)}} \\ 
\midrule
{Object} & {bowl} & {hammer} & {flashlight} & {mouse} & {duck} & {wineglass} & {scissors} & {airplane} & {stapler} & {torusmedium} \\
{Success Rate} & 100\% & 98.5\% & 95.8\% & 94.6\% & 90.4\% & 96.7\% & 89.1\% & 87.8\% & 95.5\% & 100\% \\
\midrule
{Object} & {banana} & {cylindersmall} & {waterbottle} & {watch} & {stamp} & {alarmclock} & {headphones} & {phone} & {cylindermedium} & {flute} \\
{Success Rate} & 100\% & 100\% & 100\% & 99.0\% & 100\% & 95.6\% & 97.1\% & 93.2\% & 98.3\% & 95.8\% \\
\midrule
{Object} & {cup} & {fryingpan} & {lightbulb} & {toothbrush} & {knife} \\
{Success Rate} & 92.6\% & 98.1\% & 100\% & 98.0\% & 93.0\% \\
\midrule
\multicolumn{11}{c}{\cellcolor{rqred}{GRAB-Test (5 Unseen Objects)}} \\ 
\midrule
{Object}  & apple &   binoculars & camera &   mug &   toothpaste & & & & &\\
{Success Rate} & 97.4\% & 65.5\% & 99.3\% & 93.3\% & 95.7\% \\
\bottomrule
\end{tabular}}
\caption{\small{Per-object success breakdown for the \name-stereo  policy in the  tabletop setting. }} 
\label{tab:app_obj}
\end{table*}


\begin{table*}[t]
\centering

\resizebox{\linewidth}{!}{%
\begin{tabular}{lcccccccccc}
\toprule
\multicolumn{11}{c}{\cellcolor{rqblue}GRAB-Train (25 Seen Objects)} \\ 
\midrule
Object & bowl & hammer & flashlight & mouse & duck & wineglass & scissors & airplane & stapler & torusmedium \\
Success Rate & 50.0\% & 62.5\% & 72.0\% & 74.5\% & 87.0\% & 77.0\% & 65.0\% & 60.5\% & 70.5\% & 80.5\% \\
\midrule
Object & banana & cylindersmall & waterbottle & watch & stamp & alarmclock & headphones & phone & cylindermedium & flute \\
Success Rate & 76.0\% & 56.0\% & 60.5\% & 69.5\% & 63.5\% & 58.5\% & 64.0\% & 66.5\% & 80.5\% & 74.5\% \\
\midrule
Object & cup & fryingpan & lightbulb & toothbrush & knife \\
Success Rate & 66.0\% & 43.5\% & 77.5\% & 42.0\% & 39.5\% \\
\midrule
\multicolumn{11}{c}{\cellcolor{rqred}GRAB-Test (5 Unseen Objects)} \\ 
\midrule
Object  & apple & binoculars & camera & mug & toothpaste \\
Success Rate & 63.0\% & 31.0\% & 62.0\% & 53.0\% & 59.0\% \\
\bottomrule
\end{tabular}}
\caption{\small{Per-object success breakdown for the \name-stereo policy in the kitchen setting.}} 
\label{tab:app_obj_kitchen}
\end{table*}

\subsection{State-space Policy}
\label{sec:app_omnigrasp}
We learn a state-space policy similar to Omnigrasp~\cite{luoomnigrasp} but with the modifications that instead of using desired object trajectory, we only provide a single target position. Also, Omnigrasp does not have the ability to specify which hand to use, nor can it put the object down on command. To enable these capabilities, we add additional phase variables to inform the policy of the current command. Specifically, the observation input for our omnigrasp policy is $$\obsomnigrasp \triangleq (\selfstate, \objt, \objr, \objlt, \refobjtn, \handinfo),$$
where $\objt \in \mathcal{R}^{3}$ object translation, $\objr \in \mathcal{R}^{6}$ object rotation, and object shape latent code $\objlt \in \mathcal{R}^{512}$ are privileged information provided by simulation. We compute the shape latent code $\objlt$ using BPS~\cite{prokudin2019efficient} with 512 randomly sampled points. The target object position $\refobjtn \in \mathcal{R}^{3}$ specifies the 3D position of where the object's centroid should be, and hand information $\handinfo \in \mathcal{R}^{512}$ provides information about which hand to use and when to grasp and put down. $\handinfo \triangleq (\handinfoleft, \handinforight, \handinfotime)$ contains the indicator variables to indicate whether to use the left $\handinfoleft \in \mathcal{R}^{1}$ or the right $\handinforight \in \mathcal{R}^{1}$ hand. The time variable \(\handinfotime \in \mathbb{R}^{10}\) encodes whether the policy should grasp or release the object within the next second, sampled at 0.1s intervals. 

We train the state-space policy using the same training procedure as our vision-guided policy and use the same architecture networks (removing the visual encoder). The humanoid motion prior is also the same as the vision-conditioned policy. The state-space policy is trained using a similar number of samples as the visual policy via multi-GPU RL training.

\subsection{Humanoid Motion Prior}
\label{sec:app_pulsex}

We use a similar training procedure to obtain our humanoid motion representation (PULSE-X) as Omnigrasp~\cite{luoomnigrasp}. First, we train a motion imitator, PHC-X~\cite{Luo2023-ft}, on the cleaned AMASS dataset augmented with hand motion. Then, we distill the PHC-X policy into the PULSE-X CVAE using DAgger~\cite{Ross2010-cc}. 

\begin{table}
\vspace{-5mm}
\caption{\small{Imitation result on AMASS (14889 sequences).}}
\centering

\scriptsize
\resizebox{\linewidth}{!}{%
\begin{tabular}{l|rrrrrr}
\toprule
\multicolumn{1}{c}{} & \multicolumn{5}{c}{AMASS-Train}
\\ 
\midrule
Method  & $\text{Succ} \uparrow$ & $E_\text{g-mpjpe}  \downarrow$ &  $E_\text{mpjpe} \downarrow $ &  $\text{E}_{\text{acc}} \downarrow$  & $\text{E}_{\text{vel}} \downarrow$     \\ \midrule

$\namephcx-\text{IsaacGym}$ & {99.9 \%} & {29.4} & {31.0} & {4.1} & {5.1} \\
$\namepulsex-\text{IsaacGym}$ & {99.5 \%} & {42.9} & {46.4} & {4.6} & {6.7} \\

\midrule
$\namephcx-\text{IsaacLab}$ & {99.9 \%} & {25.4} & {26.7} & {4.8} & {5.2} \\
$\namepulsex-\text{IsaacLab}$ & {99.6 \%} & {29.1} & {30.1} & {4.2} & {5.1} \\

\midrule


\bottomrule 
\end{tabular}}

\label{tab:app_im}
\end{table}

\paragraph{Motion Imitator, PHC-X} The task of motion imitation involves training a control policy to follow per-frame reference motion given input. For the motion imitation task, the input is defined as $\obsim \triangleq (\selfstate, \goalstateimitate)$, consisting of the proprioception $\selfstate$ and imitation goal $\goalstateimitate$ (one frame difference between current pose and reference motion. We follow PHC 's~\cite {Luo2023-ft} state, reward, and policy design to train one policy to imitate all motion sequences from the AMASS dataset. To increase the dexterity of the policy, we follow Omnigrasp~\cite{luoomnigrasp} and augment the AMASS dataset's whole body motion with randomly selected hand motion from the Interhand~\cite{moon2023reinterhand} and GRAB~\cite{taheri2020grab} dataset. To transition from IsaacGym~\cite{Makoviychuk2021-sx} to IsaacSim (the underlying simulation engine for IsaacLab~\cite{mittal2023orbit}), we implement the PHC-X policy in IsaacLab and train it for 5 days across using 8 GPUs. We report success rate, mean per joint position error $\gmpjpe$ (mm), local joint position error $\mpjpe$ (mm), acceleration error $\acc$ (mm/frame$^2$), and velocity error $\vel$ (mm/frame) to evaluate the performance of our policies. All metrics are completed between the simulated humanoid motion and the reference motion. From~\Cref{tab:app_im}, we can see that our IsaacLab implementation achieves comparable motion imitation results to those in IsaacGym.

\paragraph{Humanoid Motion Prior, PULSE-X} PHC-X learns the motor skills required to perform most of the actions in the AMASS dataset, and then PULSE-X learns a CVAE-like motion latent space. Specifically, PULSE learns the encoder $\encoder$, decoder $\decoder$, and prior $\prior$ to compress motor skills into a latent representation. The encoder $\encoder(\latentt | \selfstate, \goalstateimitate) \sim \mathcal{N}(\latentt | \encodermu, \encodersigma)$ computes the latent code distribution based on current input states. The decoder $\decoder(\action|\selfstate, \latentt)$  produces action (joint actuation) based on the latent code $\latentt$. The prior $\prior(\latentt | \selfstate) \sim \mathcal{N}(\latentt | \priormu, \priorsigma)$ defines a Gaussian distribution based on proprioception and replaces the unit Gaussian distribution used in VAEs \cite{Kingma2014-vh}. The prior increases the expressiveness of the latent space and guides downstream task learning by forming a residual action space. 

After the encoder, decoder, and prior is trained, similar to downstream task policies in PULSE,  we form the action space of $\policy$ as the residual action with respect to prior's mean $\priormu$ and compute the PD target $\action$: 
\begin{equation}
\label{eq:additive_action}
    \action = \decoder(\policy(\latentttask| \selfstate, \goalstate) + \priormu, \selfstate), 
\end{equation}
where $\priormu$ is computed by the prior $\prior(\latentt | \selfstate)$. The policy $\policy$ computes $\latentttask \in \mathcal{R}^{48}$ instead of the target $\action \in \mathcal{R}^{51\times 3}$ directly, and leverages the latent motion representation of $\namepulsex$ to produce human-like motion. ~\Cref{tab:app_im} shows that our implementation achieves a high success rate on the AMASS dataset.

\begin{figure*}[t!]
    \centering
    \begin{subfigure}[t]{0.48\linewidth}
        \centering
        \includegraphics[width=\textwidth]{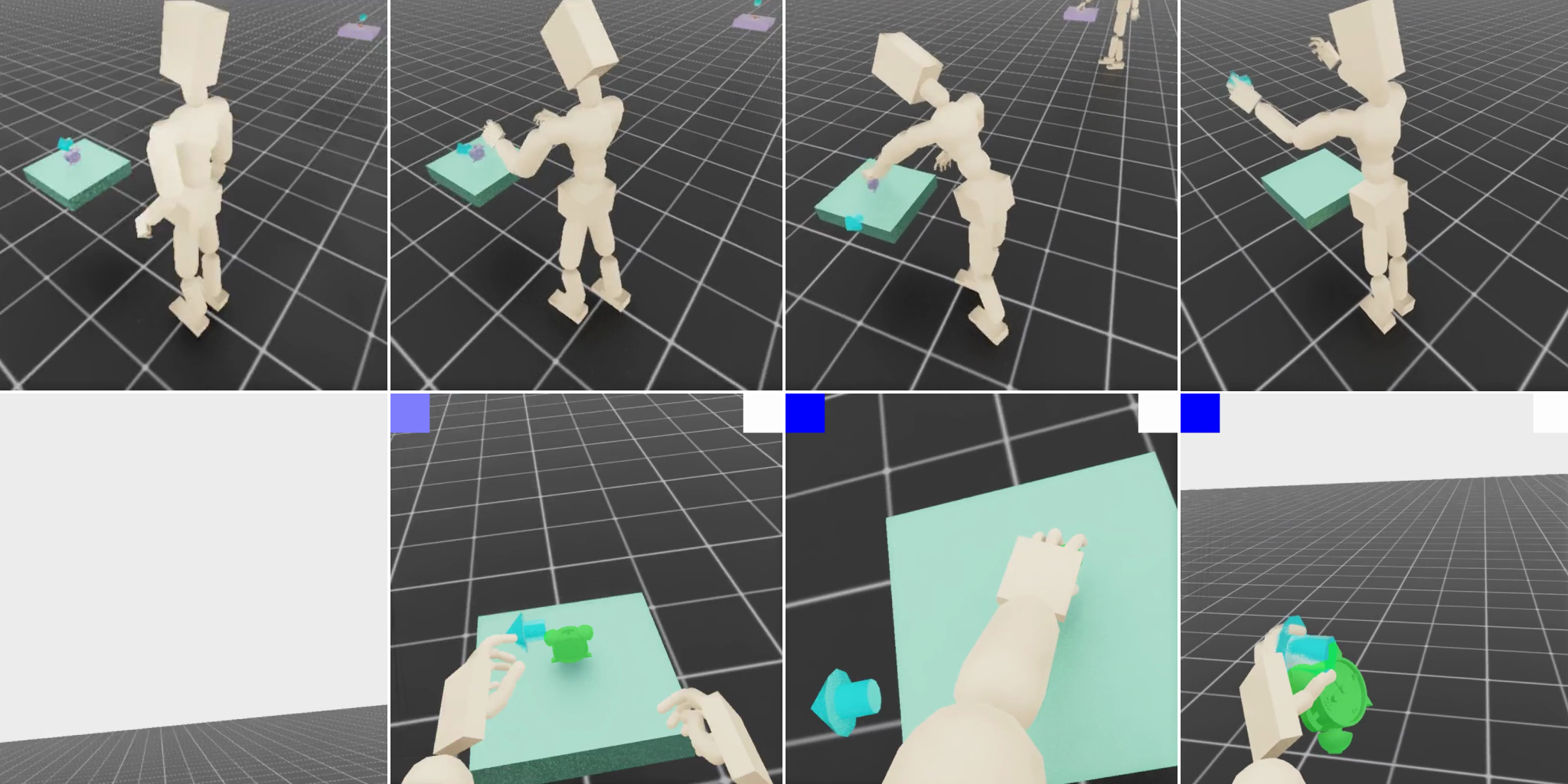}
        \caption{Left hand: Object transport. 3$^{rd}$ person and agent's camera view.}
    \end{subfigure}%
    ~ 
    \begin{subfigure}[t]{0.48\linewidth}
        \centering
        \includegraphics[width=\textwidth]{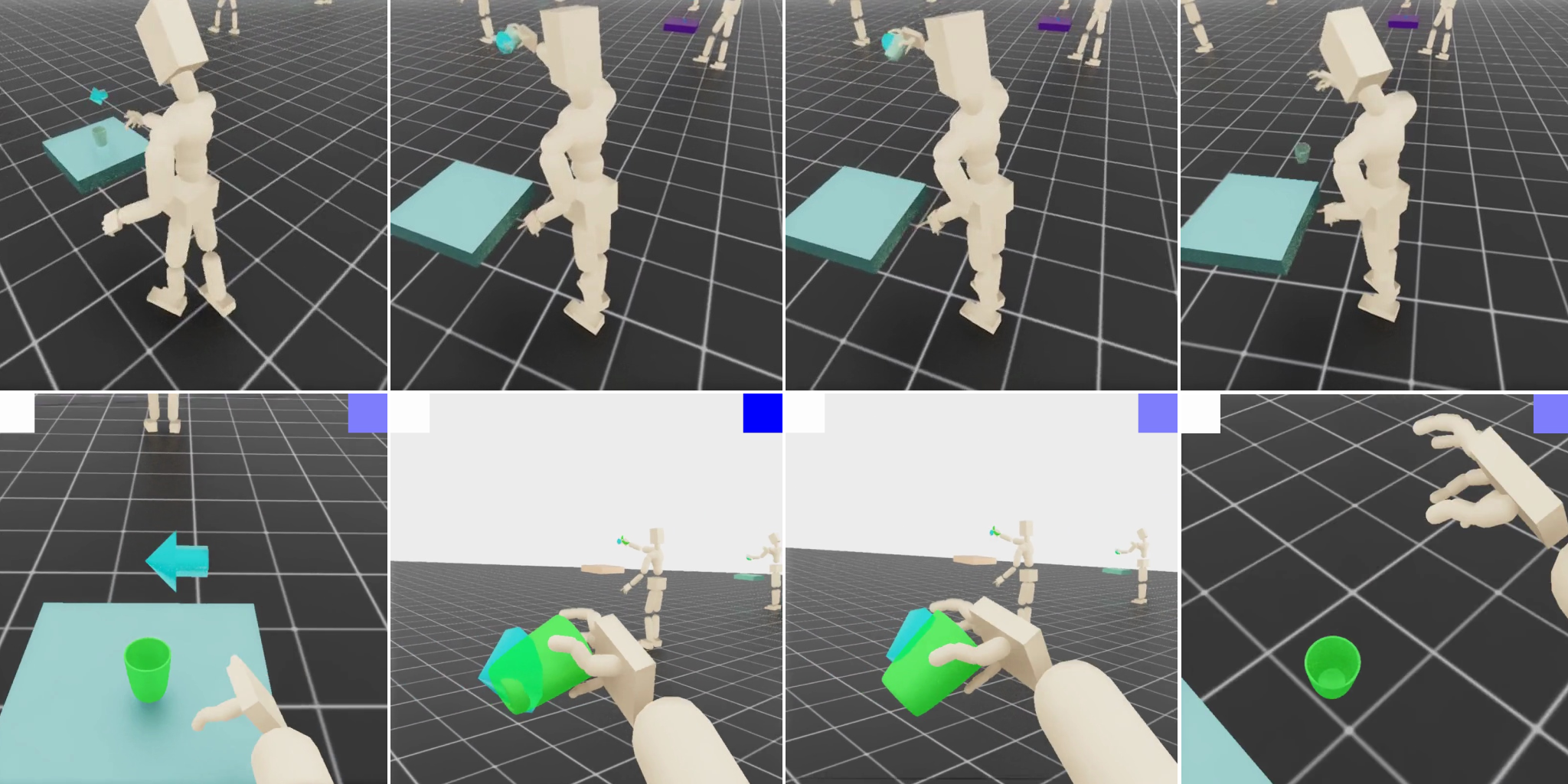}
        \caption{Right hand: Transport and release.}
    \end{subfigure} \\
    \begin{subfigure}[t]{\linewidth}
        \centering
        \includegraphics[width=0.48\textwidth]{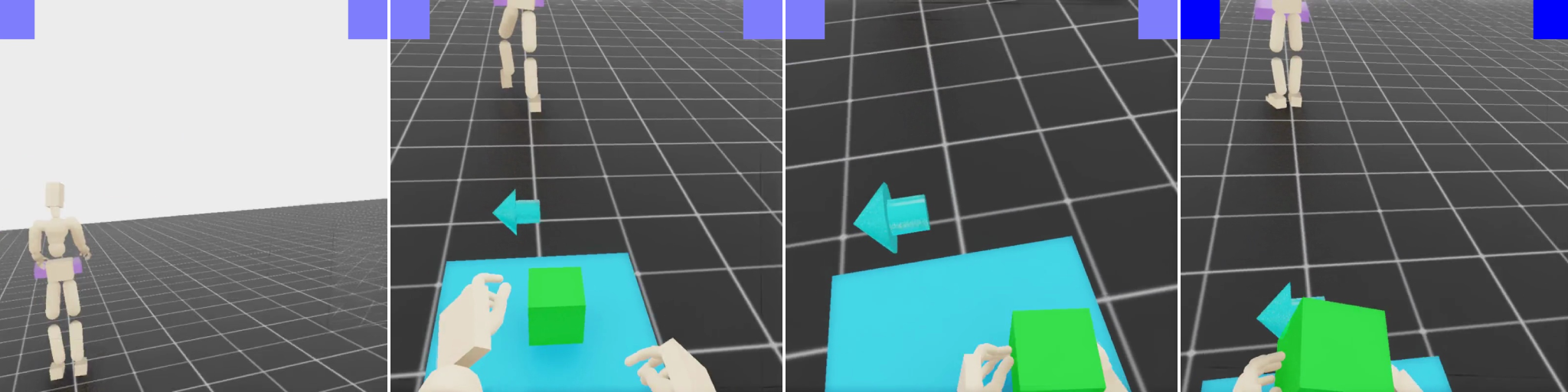}
        ~
        \includegraphics[width=0.48\textwidth]{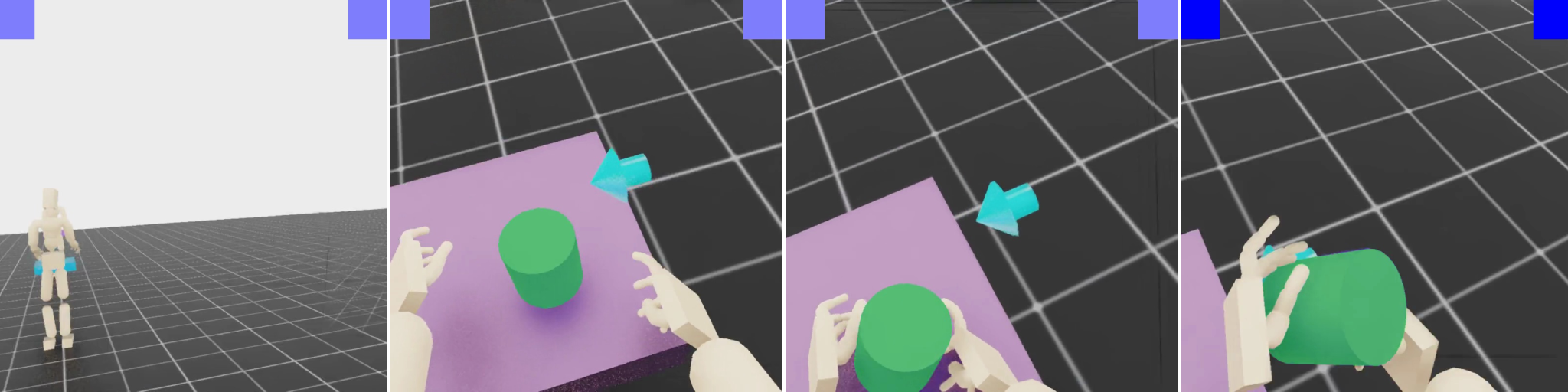}
        \caption{Both hands: Picking up large objects. Agent's camera view.}
    \end{subfigure}
    \caption{\small  \textbf{Tabletop:} \name is instructed directly from visual signals. A visual (top left and/or right) indicates in {\color{Periwinkle}purple} whether the corresponding hand should be prepared for contact. Changing to {\color{Blue}dark-blue} indicates contact should be made. Instructing the agent to use both hands enables it to transport larger objects. Changing the indicator back to {\color{Periwinkle}purple} instructs the agent to release the object.}
    \label{fig: tabletop_2}
\end{figure*}

\begin{figure*}[t!]
    \centering
    \begin{subfigure}[t]{0.98\linewidth}
        \centering
        \includegraphics[width=\textwidth]{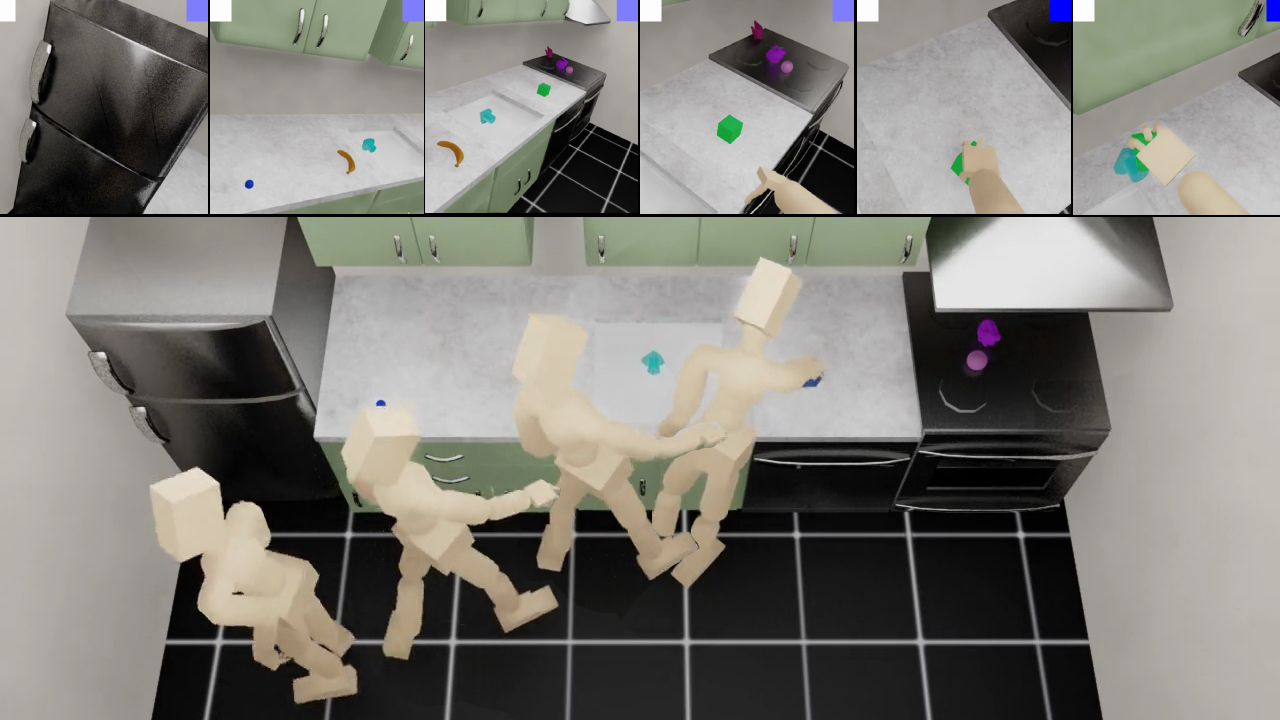}
        \caption{Object transport.}
    \end{subfigure} \\
    \begin{subfigure}[t]{0.98\linewidth}
        \centering
        \includegraphics[width=\textwidth]{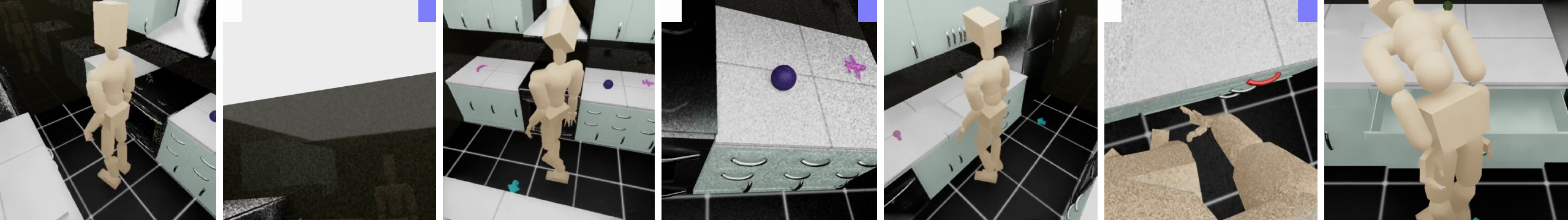}
        \caption{Drawer opening.}
    \end{subfigure}
    \caption{\small  \textbf{Emergent search in the kitchen task:} By training the agent in diverse and complex scenes, it learns generalizable behaviors and avoids overfitting. We observe that behaviors such as searching emerge. When the object is not in view, the agent scans the counter top and top drawers in order to learn of its objective and execute on it.}
    \label{fig: kitchen_task_2}
\end{figure*}

\subsection{Details about \name}
\label{sec:app_pdc}

\begin{table}[t]
\centering

\label{tab:app_hyper} 
\resizebox{\linewidth}{!}{%
\begin{tabular}{lcccccc}
\toprule
   Method & \# Envs & Learning Rate  &  $\sigma$  & $\gamma$ & $\epsilon$ & \# of samples \\ 
\midrule
$\namephcx$   & 3072 $\times$ 8 & $2\times 10^{-5}$ &  0.05  & 0.99 &  0.2  & $\sim 10^{10}$ \\
$\namepulsex$ & 3072 $\times$ 8 & $5\times 10^{-4}$ & --  & -- & -- & $\sim 10^{9}$ \\
Omnigrasp & 2048 $\times$ 8 & $2\times 10^{-5}$  &  0.36  & 0.99 &  0.2  &  $\sim 10^{9}$ \\
$\name$ & 192 $\times$ 8 & $2\times 10^{-5}$  &  0.36  & 0.99 &  0.2  &  $\sim 10^{9}$ \\
\bottomrule 
\end{tabular}}
\caption{{\footnotesize Hyperparameters for $\name$, $\namephcx$, and $\namepulsex$. $\sigma$: fixed variance for policy. $\gamma$: discount factor. $\epsilon$: clip range for PPO.}}
\end{table}
\paragraph{Hyperparameters} Hyperparameters for training PHC-X, PULSE-X, and \name can be found in \Cref{tab:app_hyper}. We do not change the hyperparameters significantly between training the visual policy and the state-space policy, demonstrating the robustness of PPO across different tasks. 

\paragraph{Rewards}  For our reward, \begin{equation}
\scriptsize
\label{eqn:lookat}
    \bs{r}^{\name}_t = 
    \begin{cases} 
         \rewardapproach + \rewardlookat, &   \|\reftpregrasp - \simthand\|_2 > 0.2 \;  \text{and} \; t < \contactstart \\ 
        \rewardpregrasp + \rewardlookat,  & \|\reftpregrasp - \simthand\|_2 \leq 0.2 \;  \text{and} \; t < \contactstart \\ 
        \rewardobj + \rewardlookat,  &  \contactstart  \leq t < \contactend \\
        (1- \mathbf{1}_{\text{has-contact}}),  &  t \geq \contactend   \\
    \end{cases}
\end{equation}
$\contactstart$ indicates the frame in which grasping should occur, and $\contactend$ is when the frame should end and the object released. $\simthand$ indicates the hands' position.  When the object is far away from the hands ($\|\reftpregrasp - \simthand\|_2 > 0.2$) and before grasping should start, the approach reward $\rewardapproach$ is similar to a point-goal \cite{Won2022-jy, Luo2023-ft} reward  $\rewardapproach = \|\reftpregrasp - \simthand\|_2 - \|\reftpregrasp - \simtphand\|_2,$, where the policy is encouraged to get close to the pre-grasp. After the hands are close enough ($\leq$ 0.2m), we use a more precise hand imitation reward: $\rewardpregrasp = w_{\text{hp}}e^{-100\|\reftpregrasp - \simthand\|_2 \times \mathbbm{1}\{\|\reftpregrasp - \refobjt\|_2 \leq 0.2 \}  } + w_{\text{hr}}e^{-100\|\refrpregrasp - \simrhand\|_2}, $ to encourage the hands to be close to pre-grasps. After the grasp start time, we switch to the object 3D location reward $\rewardobj =  e^{-5\|\targetobj - \objt \|_2} \times \mathbf{1}_{\text{correct-hand}} $ to encourage the object being moved to specific locations. $\mathbf{1}_{\text{correct-hand}}$ is the indicator random variable to determine if the correct hand has contact with the object. After the grasping should end ($t \geq \contactend$), we encourage the agent not to be in contact with the object: $\mathbf{1}_{\text{has-contact}}$ determines if any hand has contact with the object. The drawer open reward $\rewarddrawer$ is defined as how many degrees (angles defined by IsaacLab) the drawer is opened, clipped to 0 and 1.

\paragraph{Network Architecture } For our networks, we uses a two-layer CNN with each with 32 channels. Our CNN produces a latent feature space of $\mathcal{R}^{128}$. We use 6-layer MLPs of units (2048, 2048, 1024, 1024, 512, 512) with silu activation~\cite{Elfwing2017-is}. Our GRU has one layer and 128 hidden units.

\subsection{Details about Dataset} 
We use the same train/test sequence split from Omnigrasp~\cite{luoomnigrasp} and Braun \etal~\cite{braun2023physically}. For training, we use a subset of the objects and pick the ones that are more common in the households. Specifically, we use the following 25 categories:

\begin{center}
\begin{tabular}{l l l}
\texttt{torusmedium} & \texttt{flashlight} & \texttt{bowl} \\
\texttt{lightbulb} & \texttt{alarmclock} & \texttt{hammer} \\
\texttt{scissors} & \texttt{cylindermedium} & \texttt{stapler}  \\
\texttt{phone} & \texttt{duck} & \texttt{airplane} \\
\texttt{knife} & \texttt{cup} & \texttt{wineglass}  \\
\texttt{fryingpan} & \texttt{cylindersmall} & \texttt{waterbottle}  \\
\texttt{banana} & \texttt{mouse}  & \texttt{flute} \\
\texttt{headphones} & \texttt{stamp}  & \texttt{toothbrush} \\
\texttt{watch} & \\
\end{tabular}
\end{center}

Out of the 1016 training sequences from GRAB, 571 is picked. For testing, we use the same 140 sequences and object types are \texttt{apple}, \texttt{binoculars}, \texttt{camera}, \texttt{mug}, \texttt{toothpaste}.

\section{Supplementary Results}
\label{sec:app_res}

\subsection{Per-object Success Rate}

In \Cref{tab:app_obj} and \Cref{tab:app_obj_kitchen}, we report the per-object success rate for our stereo policy in the tabletop and kitchen scenes on the grab dataset. From the result, we can see that objects such as a bowl and a water bottle are easy to grasp, while objects with irregular shapes, like airplanes, are harder. Small and thin objects like a knife are also hard to pick up and have one of the lowest success rates across the kitchen and tabletop scene. Test objects such as binoculars are harder since they are bigger and difficult to grasp with one hand. The kitchen grasping paints a similar picture. No object has a zero success rate. 

\subsection{Additional Qualitative Results}

In \Cref{fig: tabletop_2} and \Cref{fig: kitchen_task_2} we present additional qualitative examples of the resulting controllers. They showcase the ability of the agent to handle objects in diverse scenes, using on-screen indicators as instructions. By training directly from vision within complex and diverse tasks, we observe search and dexterity emerges.

\end{document}